\documentclass[10pt,final,journal,twocolumn]{IEEEtran}
\usepackage[caption=false,font=footnotesize]{subfig}
\usepackage[boxed,linesnumbered,algoruled,noline]{algorithm2e}
\usepackage{graphicx}
\usepackage{amsmath,amsthm,amssymb}
\usepackage{wasysym}
\usepackage{bbm,dsfont}
\usepackage{hyperref}
\usepackage{multirow}
\usepackage{booktabs}
\usepackage{url}
\usepackage{color}

\newtheorem{definition}{Definition}

\newcommand{\etal}{et al.\thinspace}

\newcommand{\eq}[1]{Eq.\thinspace\eqref{#1}}
\newcommand{\fig}[1]{Fig.\thinspace\ref{#1}}

\newcommand{\card}[1]{\left|#1\right|}

\newcommand{\neuron}{\operatorname{neuron}}
\newcommand{\sos}{\textsc{sum-of-sum}}
\newcommand{\som}{\textsc{sum-of-max}}

\title{Improving Sparse Associative Memories by Escaping from Bogus Fixed Points}
\author{
    \IEEEauthorblockN{Zhe~Yao\IEEEauthorrefmark{1}, Vincent~Gripon\IEEEauthorrefmark{2} and Michael~G.~Rabbat\IEEEauthorrefmark{1}}
    \thanks{\IEEEauthorrefmark{1}~Z.~Yao and M.G.~Rabbat are with the Department of Electrical and Computer Engineering, McGill University, Montr\'{e}al, QC, Canada. Email: \href{mailto:zhe.yao@mail.mcgill.ca}{zhe.yao@mail.mcgill.ca}, \href{mailto:michael.rabbat@mcgill.ca}{michael.rabbat@mcgill.ca}.
    }
    \thanks{\IEEEauthorrefmark{2}~V.~Gripon is with the Electronics Department, T\'{e}l\'{e}com Bretagne, Brest, France. Email: \href{mailto:vincent.gripon@telecom-bretagne.eu}{vincent.gripon@telecom-bretagne.eu}.
    }
}

\begin{document}
\maketitle
\begin{abstract}
The Gripon-Berrou neural network (GBNN) is a recently invented recurrent neural network embracing a LDPC-like sparse encoding setup which makes it extremely resilient to noise and errors.
A natural use of GBNN is as an associative memory.
There are two activation rules for the neuron dynamics, namely \sos{} and \som{}.
The latter outperforms the former in terms of retrieval rate by a huge margin.
In prior discussions and experiments, it is believed that although \sos{} may lead the network to oscillate, \som{} always converges to an ensemble of neuron cliques corresponding to previously stored patterns.
However, this is not entirely correct.
In fact, \som{} often converges to bogus fixed points where the ensemble only comprises a small subset of the converged state.
By taking advantage of this overlooked fact, we can greatly improve the retrieval rate.
We discuss this particular issue and propose a number of heuristics to push \som{} beyond these bogus fixed points.
To tackle the problem directly and completely, a novel post-processing algorithm is also developed and customized to the structure of GBNN.
Experimental results show that the new algorithm achieves a huge performance boost in terms of both retrieval rate and run-time, compared to the standard \som{} and all the other heuristics.
\end{abstract}


\section{Introduction\label{sec:introduction}}
\subsection{Background}
Associative memories are different from conventional memory systems in that they do not require explicit addresses to the information we are interested in.
They store paired patterns.
When an associative memory is given an input pattern as the probe, the content of the input itself addresses the paired output pattern directly.
The parallel nature of the associative memory and its ability to perform pattern queries efficiently makes it suitable in a variety of application domains.
For instance, in communication networks~\cite{kaxiras2005ipstash}, routers need to determine quickly the destination port of an incoming data frame based on IP addresses.
In signal and image processing~\cite{valle2009class}, one often needs to match a incomplete or noisy version of the information with predefined templates.
Database engines~\cite{lin1976rares}, anomaly detection systems~\cite{bu2004camnids}, compression algorithms~\cite{lin2000camlz}, face recognition systems~\cite{zhang2005gabor} and many other machine learning tasks are all legitimate users of associated memories.

Among all different architectures to implement associative memories, the neural network is the most popular and widely adopted approach.
Associative memories provide two operations: storing and retrieving (also known as decoding).
In the storing operation, pairs of patterns are fed into the network, modifying the internal connections between neurons.
An aggregated representation of the patterns stored thus far is obtained.
In the retrieving operation, probes (input) are presented, which might be a noisy or incomplete version of some previously stored pattern, and the associative memory needs to retrieve the most relevant or associated pattern quickly and reliably. 

Although early exploits of this approach (e.g., linear associators~\cite{anderson1988neurocomputing,anderson1993neurocomputing2} and Willshaw networks~\cite{willshaw1969non,willshaw1971models}) date back to the 1960's, it was not until the seminal work of Hopfield~\cite{hopfield1982neural,hopfield1984neurons} in the 1980's that the neural network community started to catch on.
For the history and interesting developments on associative memories, see the recent survey~\cite{palm2013neural} and the references therein.
Quite recently, Gripon and Berrou propose a new family of sparse neural networks for associative memories~\cite{gripon2011simple,gripon2011sparse}, that we refer to as the \emph{Gripon-Berrou neural network} (GBNN), which is, in short, a variant of Willshaw networks with partite cluster structures.
GBNN resemble the model proposed by Moopenn~\etal~\cite{moopenn1987electronic}, although the motivations underneath are different.
Moopenn~\etal choose the cluster structure mainly because the resulting dilute binary codes suppress the retrieval errors due to the appearance of spurious ones in electronic circuits, whereas GBNNs use the structure to provide an immediate mapping between patterns and activated neurons, so that efficient iterative algorithms can be developed.
The cluster structure also permits GBNNs to consider ``blurred'' or ``imprecise'' inputs where some symbols are not exactly known.
In a conventional Willshaw network, a global threshold is required to determine the neuron activities, which can be either the number of active neurons in the probe, or the maximum number of signals accumulated in output neurons.
However, for GBNNs, the global threshold is no longer needed, since each cluster can naturally decide for itself.
In addition to Moopenn's model, GBNNs also allow self excitations,
as well as a new retrieval scheme \som{}~\cite{gripon2012nearly}; both help to increase the retrieval rate given incomplete patterns as probes.
This manuscript focuses on an issue with the state-of-the-art retrieval rule (\som{}) for GBNNs.
A brief description of GBNNs is given in Section~\ref{sec:gbnn}.

\subsection{Related Work}
There are three important concepts to describe the quality of an associative memory: diversity (the number of paired patterns that the network can store), capacity (the maximum amount of stored information in bits) and efficiency (the ratio between the capacity and the amount of information that the network can store when the diversity reaches its maximum).
In~\cite{gripon2011sparse}, Gripon and Berrou have shown that, given the same amount of storage, GBNN outperforms the conventional Hopfield network in all of them, while decreasing the retrieval error rate.
The initial retrieval rule used in~\cite{gripon2011sparse} was \sos{}.
Later in~\cite{gripon2012nearly}, the same authors also interpret GBNN using the formalism of error correcting codes, and propose a second retrieval rule \som{} which further decreases the error rate.
We will discuss the mechanics of both rules in Section~\ref{sec:gbnn}.
Jiang~\etal~\cite{jiang2012learning} modify GBNN to learn long sequences by incorporating directed links.
Aliabadi~\etal~\cite{aliabadi2012sparse} extend GBNN to learn sparse messages.

Another line of research focuses on efficient implementations of GBNNs.
Jarollahi~\etal~demonstrate a proof-of-concept implementation of \sos{} using \emph{field programmable gate array} (FPGA) in~\cite{jarollahi2012architecture}, though the network size is constrained to 400 neurons due to hardware limitations.
The same authors implement \som{} in~\cite{jarollahi2013reduced} and runs 1.9$\times$ faster than~\cite{jarollahi2012architecture}, since bitwise operations are used in place of a resource-demanding module required by \sos{}.
In~\cite{jarollahi2013lowpower}, the same group of authors also develop a content addressable memory using GBNNs which saves 90\% of the energy consumption.
Larras~\etal~\cite{larras2013analog} implement an analog version of the network which consumes $1165\times$ less energy but is $2\times$ more efficient both in the surface of the circuit and speed, compared with an equivalent digital circuit.
However, the network size is further constrained to $208$ neurons in total.
After analyzing the convergence and computation properties of both \sos{} and \som{}, Yao~\etal~\cite{yao2013gpugbnn} propose a hybrid scheme and successfully implement GBNNs on a GPU.
An acceleration of 900$\times$ is witnessed without any loss of accuracy.

\subsection{Contributions}
The state-of-the-art activation rule for GBNN is \som{}, outperforming \sos{} in terms of successful retrieval rate by a large margin given incomplete pattern probes; see~\cite{gripon2012nearly,yao2013gpugbnn}.
In prior discussions and experiments, it is believed that \som{} always converges to an ensemble of neuron cliques corresponding to previously stored patterns.
Lemma 3 in~\cite{yao2013gpugbnn} proves that the ensemble always exists in the final converged state.
It is also argued in~\cite{yao2013gpugbnn} that ``We can randomly choose one of them (cliques) as the reconstructed message.''
However, this interesting random selection step itself was disregarded, which in fact deserves additional attention.

The contributions of this work are three folds:
\begin{enumerate}
	\item We identify the bogus fixed point problem that the ensemble of neuron cliques only comprises a subset of the converged state where \som{} gets trapped.
	\item We propose six different heuristics, pushing \som{} beyond the bogus fixed point, which helps to improve the retrieval rate.
	\item We develop a novel post-processing algorithm which involves finding a maximum clique in the network and improves both retrieval rate and run-time.
\end{enumerate}

Although finding the maximum cliques in an arbitrary graph is a well known NP-hard problem~\cite{karp1972reducibility}, our algorithm can serve its purpose efficiently.
This is accomplished by taking into account the special structure of GBNN; see Section~\ref{sec:gbnn}.
Experimental results show that the new algorithm achieves a huge performance boost in terms of both retrieval rate and run-time, compared to the standard \som{} and all the other heuristics, which also indicates that the activation rule itself still has a plenty of room for improvements.

\subsection{Paper Organization}
The rest of this paper is structured as follows.
Section~\ref{sec:gbnn} reviews the architecture of GBNN to setup the context and explains in brief the reason that \som{} outperforms \sos{} in terms of retrieval rate.
Section~\ref{sec:problem} depicts the bogus fixed point problem after \som{} has converged.
We show that interestingly the cause of this problem is also the reason that \som{} outperforms \sos{}, and we develop a number of heuristics in Section~\ref{sec:heuristic} to push \som{} beyond the bogus fixed point.
In Section~\ref{sec:clique}, we propose the clique finding algorithm as our post-processing step to fix the problem directly and completely.
Section~\ref{sec:experiment} compares different approaches proposed in this work numerically and the paper concludes in Section~\ref{sec:summary}.

\section{Gripon-Berrou Neural Networks\label{sec:gbnn}}
\subsection{Structure}
The structure of GBNN~\cite{gripon2011simple} is closely related to the patterns it tries to store.
A pattern or a message can be viewed as a tuple of symbols.
Let us consider a message of length $C$ symbols, i.e., $(m_1, m_2, \dots, m_C)$.
Each symbol $m_c$ takes a value from a finite alphabet of size $L$, i.e., $m_c\in\lbrace x_l\vert l=1,2,\cdots,L\rbrace$.
To store messages of such a kind, we use a network of $n=CL$ neurons, which comprises $C$ clusters containing $L$ neurons each.
In this setup, cluster $c$ corresponds to the symbols $m_c$ and neuron $l$ corresponds to the specific value $x_l$ which $m_c$ takes.

GBNN is a binary valued neural network with a neuron's state being either 0 (inactive) or 1 (active).
We denote $\neuron(c,l)$ as the $l$\textsuperscript{th} neuron in cluster $c$.
Therefore, to express a message $(m_1, m_2, \dots, m_C)$ in GBNN, if $m_c=x_l$, it is equivalent to set $\neuron(c,l)=1$.
Since a symbol can only take one value at a time, for a given message, in each cluster, there is only one single active neuron accordingly.
Consequently, a message can be naturally encoded as a sparse binary string of length $n$ with exactly $C$ $1$s.
In other words, the locations of the active neurons express a particular message.
Once the values of the symbols are determined, all corresponding neurons are fixed.
These neurons form a clique (complete sub-graph), which is the representation of a message stored in a GBNN. 

\begin{figure}
\centering
\includegraphics[scale=.5]{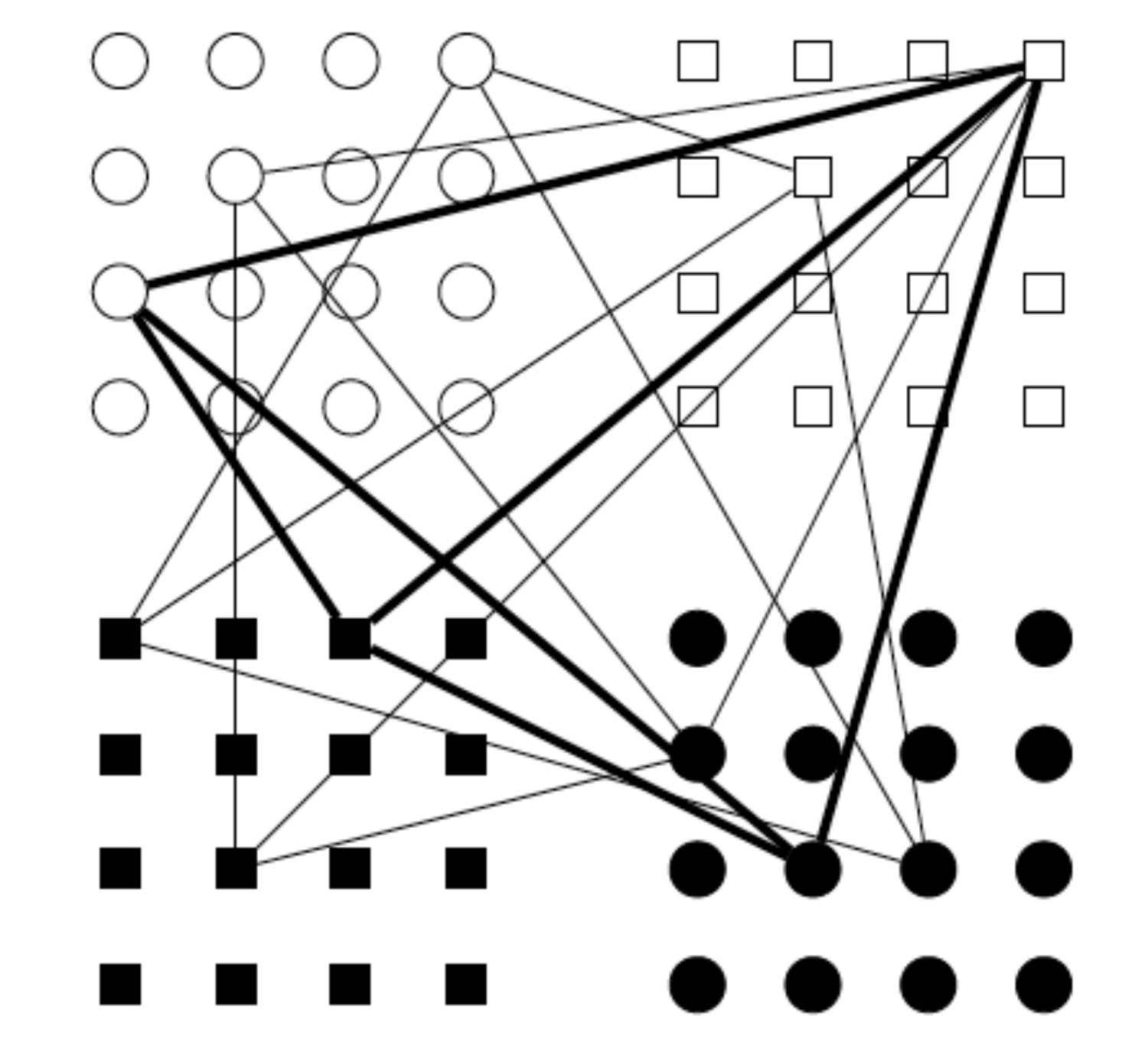}
\caption{An example of a network with $4$ clusters of $16$ neurons each~\cite{gripon2011sparse}.
We number the clusters from left to right and from top to bottom as $1 \cdots 4$.
The same scheme applies for neurons $1 \cdots 16$ within each cluster.}
\label{fig:network}
\end{figure}

\fig{fig:network} illustrates an example of GBNN with 4 clusters and each cluster has 16 neurons, with $\Circle$ for cluster 1, $\square$ for cluster 2, $\blacksquare$ for cluster 3 and $\CIRCLE$ for cluster 4.
We also number the neurons sequentially row by row from 1 to 16 in each cluster.
There are three cliques drawn in~\fig{fig:network}.
Therefore, this particular instance of GBNN stores three messages, with the black clique indicating the message (9, 4, 3, 10).

Initially, there are no edges in the network.
In the storing phase, as a message is presented, a clique with all its edges is added into the network.
As more messages are stored, more edges are added.
There will be no edge within the same cluster.
The weights on these edges can only take the value $\lbrace 1, 0\rbrace$ as well, i.e., either an edge exists or not.

For retrieval, an incomplete probe is given, e.g., $(9, ?, ?, 10)$, and the network is asked which stored message is most similar to the query.
Since the values for cluster 1 and 4 are known, to complete the query, the network activates $\neuron(1,9)$ and $\neuron(4,10)$.
Iterative activation rules can be exploited to determine which neurons in cluster 2 and 3 need to be active, \sos{}~\cite{gripon2011simple} and \som{}~\cite{gripon2012nearly} are both such examples, and we discuss them next.

\subsection{Activation Rules}
We regard active neurons to be energy sources sending out signals along the edges.
We first explain the operations of \sos{} and \som{} respectively at a high level, and then introduce rigorous notation.

\sos{} is the default activation rule for GBNN~\cite{gripon2011simple,gripon2011sparse}, which is also used in the model of Moopenn~\etal~\cite{moopenn1987electronic}.
Initially, the neurons corresponding to the remaining probes are active, transmitting signals, whereas all the neurons in the missing clusters are deactivated.
After each iteration, the neurons might receive different numbers of signals.
In each cluster, only the neurons with the most signals will remain active in the next iteration.
In contrast, \som{}~\cite{gripon2012nearly} keeps a neuron active if and only if it receives signals from every other clusters plus the self excitation.
Multiple signal contributions from the same cluster do not sum up.
However, in order for \som{} to proceed correctly, we initially activate all the neurons in the missing clusters instead, which is opposite to \sos{}.

Let $w_{(cl)(c'l')}$ denote the indicator function of whether $\neuron(c,l)$ connects to $\neuron(c'l')$, i.e.,
\begin{equation}
	\label{eq:oldindex:w}
	w_{(cl)(c'l')} =
    \begin{cases}
		1 & \mbox{$\neuron(c,l)$ connects to $\neuron(c',l')$}\\
		0 & \mbox{otherwise}.
	\end{cases}
\end{equation}
Let $v_{cl}^t$ denote the indicator function of the potential for $\neuron(c,l)$ in iteration $t$, i.e.,
\begin{equation}
    \label{eq:oldindex:v}
    v_{cl}^t =
    \begin{cases}
        1 & \quad\mbox{$\neuron(c,l)$ is active in iteration $t$}\\
        0 & \quad\mbox{otherwise}.
    \end{cases}
\end{equation}
We denote by $s_{cl}^t$ the count of the number of signals $\neuron(c,l)$ receives at iteration $t$.

The \sos{} decoding dynamics are given by
\begin{align}
    s_{cl}^t & = \gamma v_{cl}^t + \sum_{c'=1}^{C}{\sum_{l'=1}^{L}{(v_{c'l'}^t w_{(c'l')(cl)})}}\label{eq:oldindex:score}\\    s_{c,\max}^t & = \max_{1\leq l\leq L}{s_{cl}^t}\label{eq:findmax}\\
    v_{cl}^{t+1} & =
    \begin{cases}
        1 & \quad\mbox{if $s_{cl}^t = s_{c,\max}^t$}\label{eq:chooseMax}\\
        0 & \quad\mbox{otherwise},
    \end{cases}
\end{align}
where $\gamma \ge 0$ is a reinforcement factor, representing the strength of self excitations.
\som{} modifies the procedure to be
\begin{align}
    s_{cl}^t &= \gamma v_{cl}^t + \sum_{c'=1}^{C}{\max_{1\leq l'\leq L}{\left( v_{c'l'}^tw_{(c'l')(cl)}\right)}}\label{eq:newrule:score}\\
    v_{cl}^{t+1} &=
    \begin{cases}
        1 & \quad\mbox{if} \quad s_{cl}^t = \gamma + C - 1\\
        0 & \quad\mbox{otherwise}.
    \end{cases}\label{eq:newrule:select}
\end{align}
Thus for \som{}, one neuron receives at most one signal from each cluster.

This process continues until the network converges if it ever does.
In fact, \cite{yao2013gpugbnn} already provides a simple example which shows that \sos{} might oscillate and also proves that \som{} is guaranteed to converge.
This is one of the reasons we prefer \som{} over \sos{}.

\begin{figure}
  \centering
  \includegraphics[scale=.5]{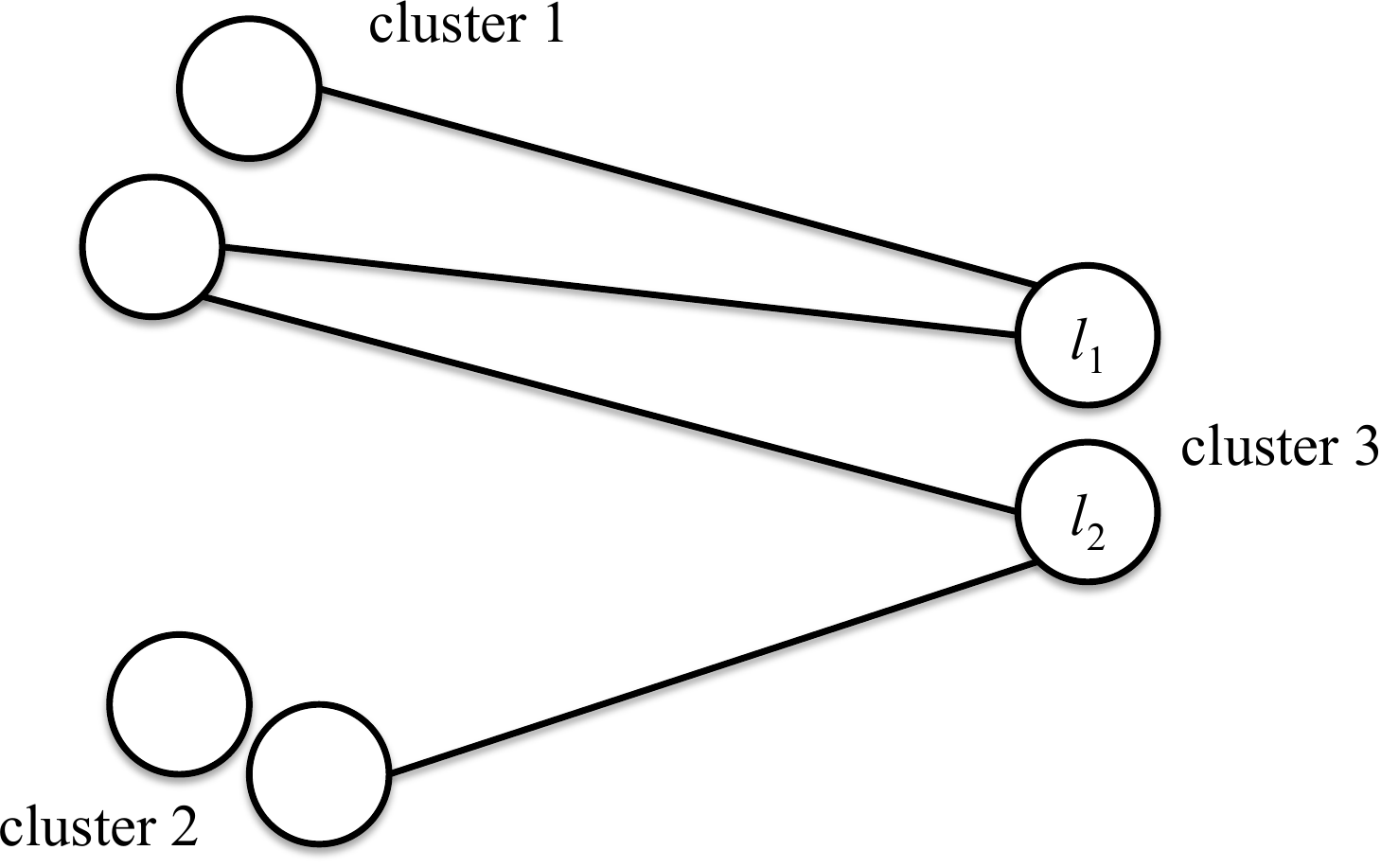}
  \caption{Illustration of the \sos{} trap. Only the signals flowing into cluster $3$ are drawn.}
  \label{fig:trap}
\end{figure}

The other reason can be illustrated by~\fig{fig:trap}, where both neurons $l_1$ and $l_2$ in cluster 3 have two individual signals.
Only the signals flowing into cluster 3 are drawn.
Neuron $l_1$ receives two signals from neurons in the same cluster (cluster 1), whereas $l_2$ receives two signals from different clusters.
In this case, $l_2$ should be favored, since we know by design, a stored message corresponds to a clique, and each cluster can only accommodate one single active neuron.
\sos{} activates both $l_1$ and $l_2$ since they have equal number of signals received, whereas \som{} can differentiate between them and favor $l_2$ as desired.
A worse but possible situation is that $l_1$ receives more signals than $l_2$.
In this case, only $l_1$ will remain active, and the correct $l_2$ is deactivated.
Therefore, for \sos{}, the decoding errors may propagate from the current iteration to the next.

In summary, \sos{} counts individual signals possibly propagating decoding errors, whereas \som{} counts cluster-wise contributions with the desired decoding procedure preserved.
This is exactly the reason that \som{} outperforms \sos{} in terms of retrieval rate by a large margin, especially in challenging scenarios, e.g., either the number of stored messages or the number of erased symbols increases.
For detailed performance comparisons and different initialization schemes, see~\cite{gripon2012nearly,yao2013gpugbnn}.

\section{Bogus Fixed Point Problem\label{sec:problem}}
Concerning the activation rules, all prior discussions and experiments concentrate on comparative studies between \sos{} and \som{}.
However, little effort has been put into investigations of the activation rules themselves.
In this section, we will illustrate a formerly disregarded aspect and identify a hidden issue embedded in \som{}.
 
\begin{figure}
\centering
\includegraphics[scale=.5]{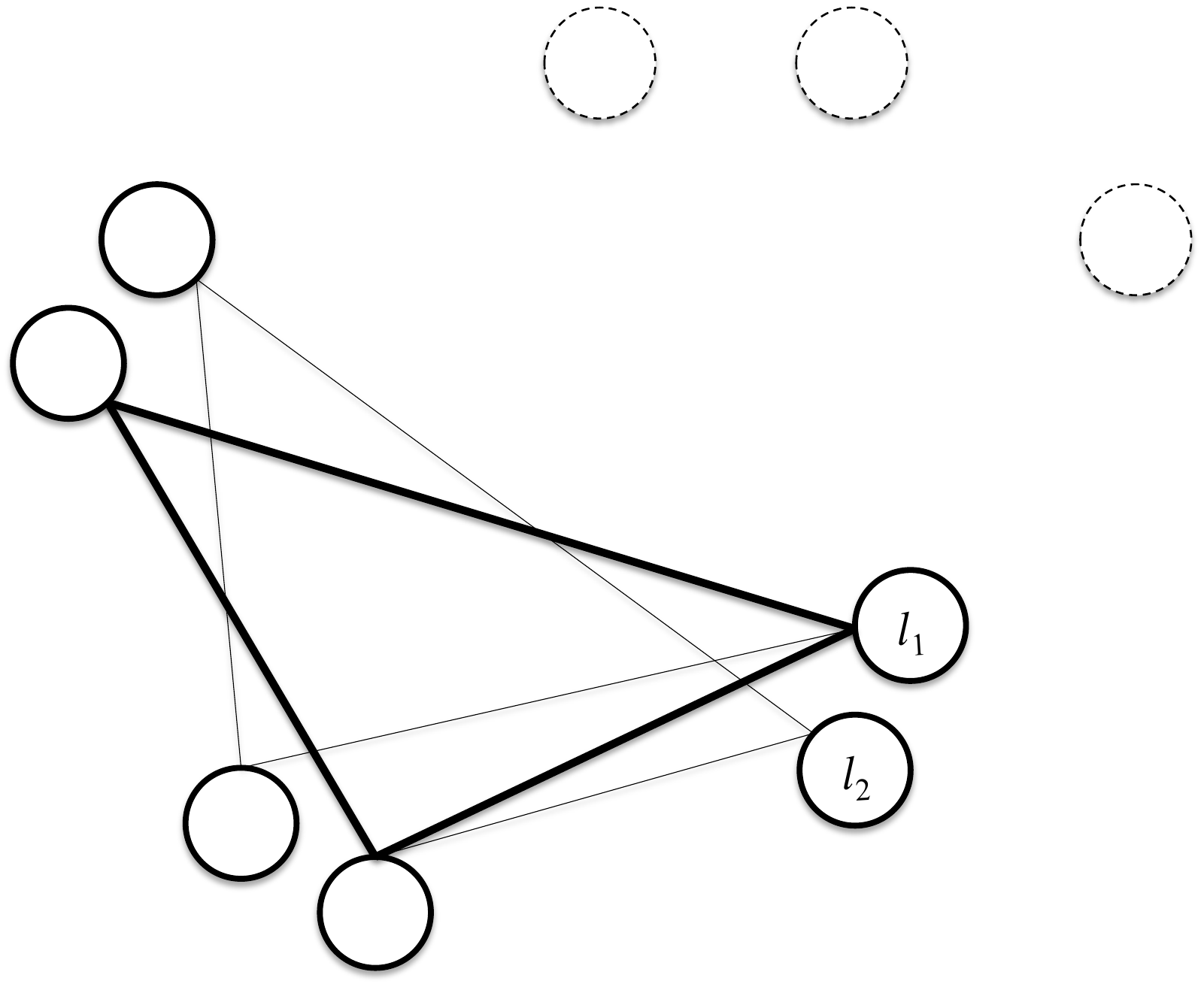}
\caption{Illustration of the overlooked \som{} problem after the network has converged. Only some effective signal paths are drawn. Dashed circles are active neurons in other clusters, they all have signal contributions to each of the sold neurons on the left corner.}
\label{fig:problem}
\end{figure}

\fig{fig:problem} depicts a part of the final state after \som{} has converged.
For brevity and clearness, we do not draw the edges of the network described by $w_{(cl)(c'l')}$, but only the signal paths.
For the same purpose, we also omit a large number of signal paths, where each dashed circle is some active neuron in a different cluster, contributing signals to each of the solid neurons.

Let us focus on the solid neurons on the left corner.
There are three clusters and each of them has two active neurons in the converged state.
All six neurons receive signals from every other cluster, including those signals from the dashed neurons which we do not draw in~\fig{fig:problem}.
Hence, \som{} will keep them all active.
To decode a message, we try to find a $C$-partite clique, since we have $C$ clusters.
The problem is that after \som{} has converged, other signal paths which do not form a clique also exist, e.g., the thinner lines in~\fig{fig:problem}.
A subset of a clique is a clique of a smaller size.
Therefore, if the message is decoded correctly, a clique of size three, the only dark triangle we can find, needs to be identified.
This is simply a small illustration of a three-cluster scenario. 
Imagine the complication when all the signal paths are presented and the network gets larger.
If we do not select active neurons strategically but arbitrarily pick some random neurons in the converged state as the final answer, it is highly likely that we will not choose the ones that can actually form a clique.

One might wonder, since each stored message corresponds to a clique, how it is possible for these thinner lines which cannot form a clique to remain in the converged state.
For a neuron to receive a signal from others, two prerequisites exist:
\begin{enumerate}
\item There is an edge. This part is fixed after the storing phase finishes.
\item The neuron on the other side of the edge is active, i.e., not all edges are active. This part is dynamic as the retrieval process continues.
\end{enumerate}
The storing phase preserves all cliques, whereas the retrieving phase only searches for cliques with active edges (edges with signals).
As a matter of fact, these thinner lines are part of the cliques which corresponds to other messages (we do not draw them in~\fig{fig:problem}), but they fail to form an clique as the decoding result, which requires all of its edges to be active.
However, \som{} preserves all these thinner lines undesirably.

\section{Heuristic Solutions\label{sec:heuristic}}
In Section~\ref{sec:problem}, we have shown the bogus fixed point problem of \som{}, that the cliques corresponding to stored messages are hidden in the converged state.
Interestingly, the cause of the problem is exactly the reason that \som{} outperforms \sos{} in terms of retrieval rate: it does not differentiate individual signals from the same cluster.
For instance, in~\fig{fig:problem}, both neuron $l_1$ and $l_2$ receive contributions from the other two clusters, hence at the end they both remain active.
We need some post-processing to break the tie and let \som{} continue with its work.

We choose to deactivate some neurons when \som{} gets trapped and does not make further improvements until a clique of size $C$ is hopefully found.
Since it is the cluster-wise signal contributions that lead \som{} into the bogus fixed point, we take individual signals into account to fix the problem.
For example in~\fig{fig:problem}, neuron $l_1$ receives three individual signals while neuron $l_2$ receives only two.
In this case, neuron $l_1$ should be favored, because more individual signals means that more stored messages have a symbol corresponding to neuron $l_1$, which also means that we have a larger chance to successfully find a clique using this neuron.

Two options are available: either we keep the neuron with the most individual signals active, or we turn the neuron with the fewest individual signals deactivated.
Once the tie is broken, \som{} is able to continue eliminating neurons.
In this particular example, both options will favor neuron $l_1$ over $l_2$, and irrelevant neurons in other clusters will all be deactivated in the next iteration, with the desired dark triangle preserved.

Up to now, we are discussing the operations within a single cluster.
The next question is in which cluster we should break the tie.
\fig{fig:problem} is again a simple illustration in the sense that all three clusters of interest have exactly the same number of active neurons.
In general, the bogus fixed point will have clusters with different numbers of active neurons.
We can choose the cluster with either the most or the fewest active neurons to start off.

Therefore, we consider the following four heuristics:
\begin{itemize}
\item activating the neuron with the \textsc{m}ost individual signals in the cluster with the \textsc{m}ost active neurons, (\textsc{mm}).
\item activating the neuron with the \textsc{m}ost individual signals in the cluster with the \textsc{f}ewest active neurons, (\textsc{mf}).
\item deactivating the neuron with the \textsc{f}ewest individual signals in the cluster with the \textsc{m}ost active neurons, (\textsc{fm}).
\item deactivating the neuron with the \textsc{f}ewest individual signals in the cluster with the \textsc{f}ewest active neurons, (\textsc{ff}).
\end{itemize}
We argue at this point that intuitively \textsc{fm} and \textsc{ff} ought to outperform \textsc{mm} and \textsc{mf} in terms of retrieval rate.
\textsc{mm} and \textsc{mf} activate a particular neuron in some cluster, and they tend to take a guess too early in the retrieving process, whereas \textsc{fm} and \textsc{ff} eliminate unlikely neurons and let \som{} clean up the path in the coming iterations.
In terms of run-time, the competition should be reversed due to the same reasoning.
We also argue that \textsc{mf} ought to outperform \textsc{mm}, since both of them are required to choose a neuron in some cluster first and then presume that the selected neuron will be in the desired clique.
If the neuron comes from a cluster with more active neurons (\textsc{mm}), we are more likely to make an erroneous guess.
Simulation results are available in Section~\ref{sec:ex1}.

We also consider two other alternatives:
\begin{itemize}
\item deactivating the neuron with the \textsc{f}ewest \textsc{e}dges across the network, i.e., the node with the fewest neighbors, (\textsc{fe}).
\item deactivating the neuron with the \textsc{f}ewest \textsc{s}ignals across the network, i.e., the node with the fewest \emph{active} neighbors, (\textsc{fs}).
\end{itemize}
These two options seem to have interesting justifications.
\textsc{fe} actually tends to lower the activation rate of neurons that appeared less often in the stored messages.
As a matter of fact, neurons with fewer edges are expected to have appeared less often in the storing process.
Therefore, \textsc{fe} intuitively takes into account the relative frequency of the corresponding symbols in stored messages.
\textsc{fs} does something similar.
However, it is only interested in the subset of the neurons which the input probe tries to address.

Note that all these heuristics aim to increase the chance of finding a correct clique, but none of them guarantees a clique to be found.
Although the heuristics do not have impact on the convergence of the retrieving process, it is possible that eventually some cluster will have all its neurons deactivated, and in the next iteration, all neurons across the network will deactivate according to~\eq{eq:newrule:score} and~\eq{eq:newrule:select}.
After applying these heuristics once, \som{} might get caught in some other state later again, thus we will have to break the tie several times along the way until either a clique is found or some cluster has all its neurons deactivated.

\begin{figure}
\centering
\includegraphics[scale=.5]{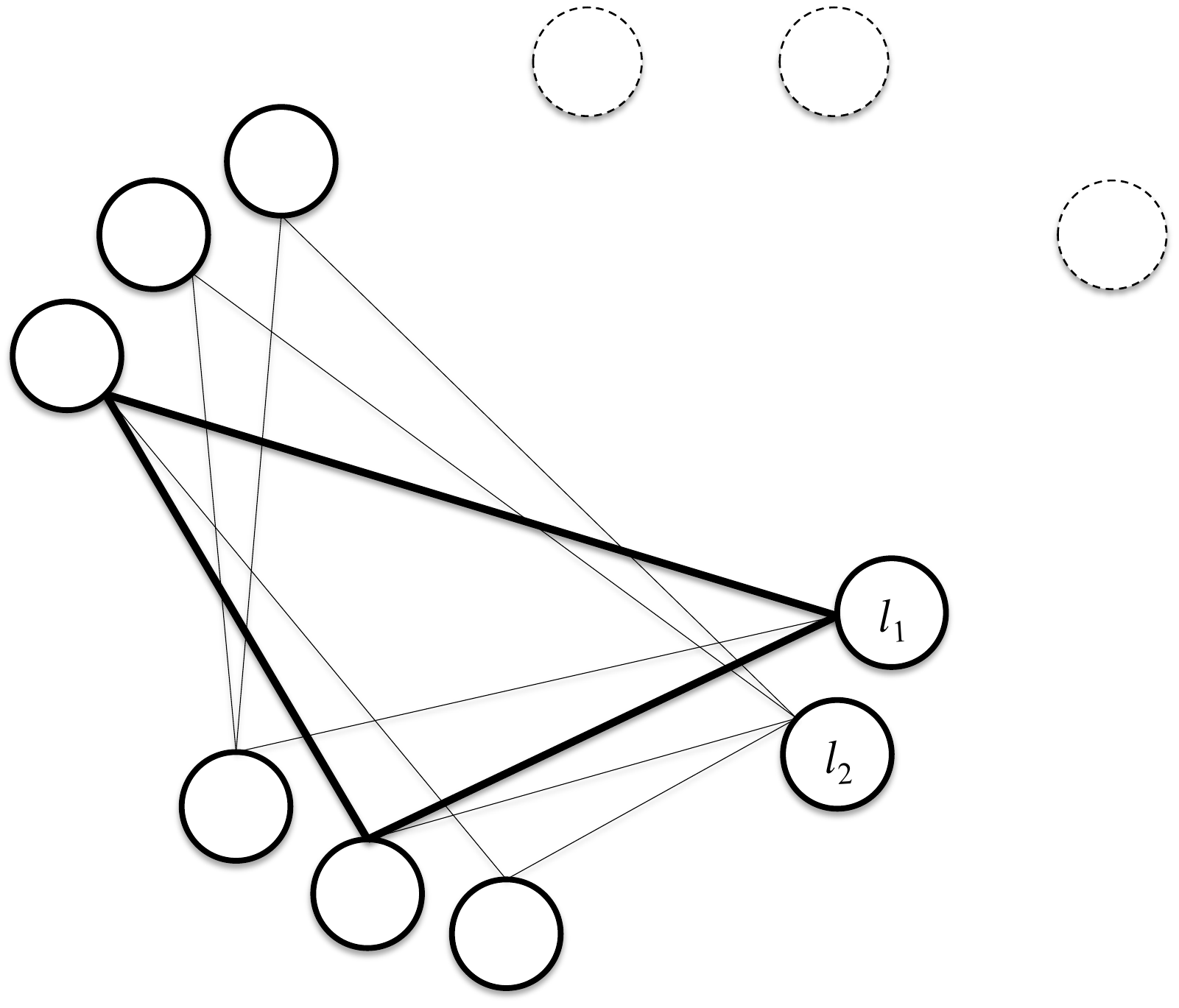}
\caption{Illustration of the failure of the heuristic \textsc{ff}. Although \textsc{ff} performs better than the other heuristics, it still cannot guarantee an active clique to be found.}
\label{fig:ff}
\end{figure}

Here we will present an example showing that these heuristics are not guaranteed to find a desired clique.
Let us assume that we use the heuristic \textsc{ff}, that is to eliminate the neuron with the fewest signals in the cluster with the fewest active neurons.
We will see numerical experiments later in Section~\ref{sec:ex1} that \textsc{ff} actually performs much better compared to the other heuristics.
However, look at~\fig{fig:ff}.
The dark triangle is again the only active clique we can find at this stage.
According to \textsc{ff}, the cluster containing $l_1$ and $l_2$ will be chosen, since it has the fewest active neurons, and $l_1$ will be eliminated right away, since it involves fewer individual signals than $l_2$.

\section{Maximum Clique\label{sec:clique}}
\begin{figure}
\centering
\includegraphics[scale=.6]{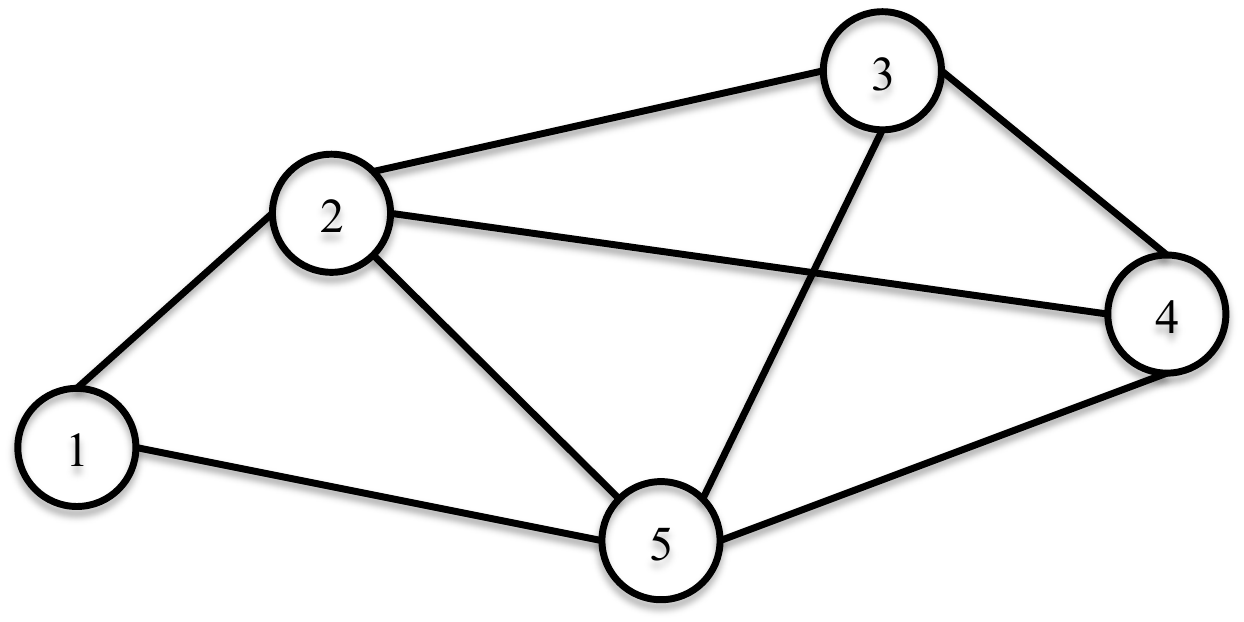}
\caption{The difference between a maximal clque and a maximum clique. There are two maximal cliques in the graph, i.e., $\lbrace1,2,5\rbrace$ and $\lbrace2,3,4,5\rbrace$, whereas the latter is a maximum clique.}
\label{fig:cliquediff}
\end{figure}

We first introduce necessary notation and definitions.
Let $G=(V,E)$ be an undirected graph, where $E$ is the set of the edges and $V=\lbrace v_1,v_2,\cdots,v_n\rbrace$ is the set of the nodes with $n$ being the number of nodes in $G$.
We denote by $N(v_i)$ the neighborhood of the node $v_i$, i.e.,
\begin{equation}
	N(v_i)=\lbrace v_j\in V\vert (v_i,v_j)\in E\rbrace.
\end{equation}
\begin{definition}
A maximal clique is a clique which is not a subgraph of any other clique in the graph $G$.
\end{definition} 
\begin{definition}
A maximum clique $\omega(G)$ is a maximal clique of the largest size in the graph $G$.
\end{definition}
To better differentiate these two definitions, see \fig{fig:cliquediff}.
The clique $\lbrace1,2,5\rbrace$ and $\lbrace2,3,4,5\rbrace$ are both maximal cliques, since adding an extra node does not form a larger clique.
The clique $\lbrace2,3,4,5\rbrace$ is a maximum clique, since it is a maximal clique of the largest size (4 in this example). 

\subsection{Motivation}
We have seen in previous sections that retrieving a message is equivalent to finding an active clique in the network.
We have also seen in Section~\ref{sec:heuristic} that although extremely helpful for \som{} to escape from the bogus fixed point, the heuristics proposed there are not guaranteed to find a clique.

Since there are no connections among the neurons within the same cluster, given a $C$-clustered GBNN, the network is a $C$-partite graph.
The active cliques we aim to find are all cliques of size $C$, and there is no clique that contains more than $C$ neurons.
Thus these cliques are also the maximum cliques in the network.
Therefore, the question becomes if we can find the maximum cliques $\omega(G)$ in a $C$-partite graph $G$ reliably and efficiently.

\subsection{Prior Work}
Finding the maximum cliques in an arbitrary graph is one of Karp's famous 21 NP-hard problems~\cite{karp1972reducibility}.
Generally speaking, to solve a difficult problem, two approaches exist: either answer it exactly but inefficiently or approximately but quickly.
However, the maximum clique problem does not lend itself to the approximate approach~\cite{williamson2011design} in the sense that for any $\epsilon > 0$, it cannot be approximated in polynomial time with the performance ratio
\begin{equation}
	\frac{\mbox{size of a maximum clique}}{\mbox{size of the approximate clique}} = O(n^\epsilon),\label{eq:polynomial}
\end{equation}
where $n$ here is the number of neurons in the network.

Modern algorithms for solving the maximum clique problem all follow the same branch-and-bound meta template.
Typical representatives are the algorithms of Carraghan and Pardalos~\cite{carraghan1990exact}, \"Osterg{\aa}rd~\cite{ostergaard2002fast}, Tomita and Seki~\cite{tomita2003efficient}, Konc and Janezic~\cite{konc2007mcqd}, and Pattabiraman~\etal~\cite{pattabiraman2012mcp}.
After presenting the classic algorithm~\cite{carraghan1990exact} as in Algorithm~\ref{alg:cp}, we will spend some time to explain the rationale behind.
The improvements by the others will then be much easier to follow in an incremental manner.

\begin{algorithm}
\DontPrintSemicolon
\SetDataSty{textit}
\SetKwFunction{clique}{clique}
\SetFuncSty{textbf}
\KwIn{An undirected graph $G=(V,E)$}
\KwOut{$Q_{max}$ (a maximum clique of $G$)}
\BlankLine\BlankLine\BlankLine\BlankLine
\textbf{global} $Q$ and $Q_{max}$\;
$Q \gets \emptyset$ and $Q_{max}\gets\emptyset$\;
\clique{$V$}\;
\Return{$Q_{max}$}
\BlankLine\BlankLine\BlankLine\BlankLine
{\bf function} \clique{$U$}:\;
\Indp\If{$\card{U}=0$ {\bf and} $\card{Q}>\card{Q_{max}}$}{
	$Q_{max}\gets Q$\;
	\Return
}
\While{$U\neq\emptyset$}{
	\If{$\card{U}+\card{Q}<\card{Q_{max}}$}{
		\Return
	}
	$v\gets U[1]$\;
	$U\gets U\setminus\lbrace v\rbrace$\;
	$Q\gets Q\cup\lbrace v\rbrace$\;
	\clique{$U\cap N(v)$}\;
	$Q\gets Q\setminus\lbrace v\rbrace$\;
}
\caption{The classic maximum clique finding algorithm by Canrraghan and Pardalos~\cite{carraghan1990exact}}
\label{alg:cp}
\end{algorithm}

Algorithm~\ref{alg:cp} starts by defining two global sets $Q$ and $Q_{max}$, where $Q_{max}$ records the largest clique we have encountered thus far, and $Q$ is the current clique we are investigating.
After initializing $Q$ and $Q_{max}$, the algorithm recursively call the function \textbf{clique}, which takes a node set $U$ (when implemented, it can be a vector or an array in memory), the current sub-graph of interest, as its argument.
At line 3, the full node set $V$ is under consideration.
Algorithm~\ref{alg:cp} checks the maximal clique containing $v_1$ first, and then that of $v_2$ and so on.
It tries to enumerate all the maximal cliques and then keep the largest one as the final answer.
When the algorithm terminates, $\omega(G)$ is reported by $Q_{max}$.

Every time the algorithm reaches lines 6--8, a maximal clique has been recorded in $Q$.
Thus, if its size exceeds the current largest clique, we replace $Q_{max}$ with $Q$.
From lines 14--18, as long as the node set is not empty, we take one node out at a time, assuming it is in the current clique under investigation, and then we recursively check a smaller sub-graph.
Since $Q$ always records a clique at any time, we have to ensure that all the nodes in the next sub-graph we are about to check need to connect to the nodes in $Q$, this is accomplished by the set intersection operation at line 17.
Using the terminology of branch-and-bound algorithms, line 17 is the branch step, which goes one level deeper into the binary search tree, whereas lines 11--13 are the bound step.
Even if all the nodes in the current set pool $U$ are added into the current clique $Q$, they still cannot form a larger clique than the largest one $Q_{max}$ having been seen, and so we prune the search branch immediately.
Here $\card{Q_{max}}$ is a lower bound on the global optimum solution $\card{\omega(G)}$, whereas $\card{U}$ can be regarded to be an upper bound on the local optimum solution $\card{\omega(U)}$.

\"Osterg{\aa}rd~\cite{ostergaard2002fast} accelerates the algorithm by some extra bookkeeping as well as reversing the ordering of investigating from $v_n$ back to $v_1$, so that a new type of prune technique can be applied.
The vertex coloring problem is a closely related NP-hard problem which can be used to accelerate the clique finding process.
In this task, one is required to assign colors to vertices in the graph such that no adjacent vertices share the same color.
Since solving the coloring problem exactly is also exponentially difficult, Tomita and Seki~\cite{tomita2003efficient} use an approximate vertex coloring algorithm in a greedy manner at line 14 to construct a relaxed upper bound and then replace $\card{U}$ at line 11, so more search branches can be cut as early as possible.
Konc and Janezic~\cite{konc2007mcqd} take a similar approach to~\cite{tomita2003efficient}, but they also reduce the steps needed for the approximate vertex coloring algorithm by carefully maintaining a non-increasing coloring order of vertices.
Since evaluating this auxiliary function also takes much time, it is not economical to recompute a new bound at every level.
Therefore, \cite{konc2007mcqd} also provides a variant of the algorithm which only carries out the approximate vertex coloring at top levels based on empirical experience, so that the total run-time will not be affected most of the time by the unnecessary computations when the time-consuming vertex coloring starts to slow down the whole algorithm.
The relatively recent work by Pattabiraman~\etal~\cite{pattabiraman2012mcp} argues that most of the real world networks are rarely dense networks.
After updating $Q_{max}$ at line 7 of Algorithm~\ref{alg:cp}, they prune the search domain in successive recursive calls by only checking the nodes with degrees larger than $\card{Q_{max}}$, since a node with a smaller degree can never be part of a clique larger than the current one found.

\subsection{Proposed Approach}
The aforementioned algorithms are all for finding the maximum cliques in general graphs, whereas GBNNs have a stringent nice structure.
Algorithms in existing literature require enumerating all the maximal cliques until every node has been investigated, otherwise there is no way to ensure the final $Q_{max}$ is the largest one.
However, as pointed out above, a GBNN is a $C$-partite graph with the active clique we try to find being a maximum clique of the exact size $C$.
In other words, we know in advance the targeted size, which can be used directly to prune the search space.
We also know that once a maximal clique has been found, it is guaranteed to be a maximum clique.
All of the extra information can be used to accelerate our approach.
We present our algorithm as in Algorithm~\ref{alg:kp}, mimicking the structure of Algorithm~\ref{alg:cp}.

\begin{algorithm}
\DontPrintSemicolon
\SetKwFunction{clique}{clique}
\SetKwFunction{update}{update}
\SetFuncSty{textbf}
\SetDataSty{textit}
\SetKwData{found}{found}
\SetKwData{level}{level}
\SetKwData{subgraph}{subgraph}

\KwIn{GBNN structure $G=(V,E)$ with $C$ clusters after \som{} has converged to the bogus fixed point}
\KwOut{$Q$ (an active clique)}
\BlankLine\BlankLine\BlankLine\BlankLine
\textbf{global} $Q$ and \found\;
$Q \gets \emptyset$ and \found$\gets$ \textbf{false}\;
obtain a smaller graph $G'(V',E')$ with $C'$ clusters by eliminating non-erased clusters and erased clusters but with only one active neuron\;
obtain an even smaller graph $\widetilde{G}(\widetilde{V},\widetilde{E})$ with $\widetilde{C}$ clusters by eliminating inactive neurons in the remaining clusters\;
split $\widetilde{V}$ into $\widetilde{C}$ sets $R=\lbrace R_0,R_1,\cdots,R_{\widetilde{C}-1}\rbrace$ with each accommodating the neurons in a different cluster\;
\clique{$R$}\;
\Return{$Q$}
\BlankLine\BlankLine\BlankLine\BlankLine
{\bf function} \clique{$U$}:\;
\Indp$\level\gets\card{Q}$\;
\If{$\level=\widetilde{C}$}{
	$\found\gets\textbf{true}$\;
	\Return
}
sort $U_{\level}$ according to the degrees of its nodes\;
\While{$U_{\level}\neq\emptyset$}{
	\If{any of $\lbrace U_{level+1}, \cdots, U_{\widetilde{C}-1}\rbrace = \emptyset$}{
		\Return
	}
	$v\gets U_{level}[1]$\;
	$U_{level}\gets U_{level}\setminus\lbrace v\rbrace$\;
	$Q\gets Q\cup\lbrace v\rbrace$\;
	$\subgraph\gets\update{U, v, \level}$\;
	\clique{\subgraph}\;
	\If{\found}{
		\Return	
	}
	$Q\gets Q\setminus\lbrace v\rbrace$\;
}
\BlankLine\BlankLine\BlankLine\BlankLine
\Indm{\bf function} \update{$U,v,\level$}:\;
\Indp\For{$i=level+1\;\textbf{to}\;\widetilde{C}-1$}{
	$U_i\gets U_i\cap N(v)$\;
}
sort $\lbrace U_{level+1}, \cdots, U_{\widetilde{C}-1}\rbrace$ according to the number of neurons in each set\;
\Return $U$

\caption{The proposed algorithm to fully exploit the nice structure of GBNN.}
\label{alg:kp}
\end{algorithm}

\subsection{Justifications}
Two reasons prevent GBNNs in our context from being a sparse network:
\begin{enumerate}
\item Although the representation of a given message in GBNN is extremely sparse, we focus on the situation when the retrieval scenario is challenging for the network.
This means when the number of stored messages is large, the network is dense,
which also means the converged state is a highly connected graph; see~\fig{fig:ff}.
\item Even if we store only a few messages, i.e., the total number of edges is small, we still get fully connected sub-graphs (cliques); meanwhile a large number of neurons are isolated.
This is not a sparse network in the usual sense either.
\end{enumerate}
These facts motivate our extra initialization at lines 3--5 of Algorithm~\ref{alg:kp} right before the main steps of the algorithm take place.
Non-erased clusters and erased clusters but with only one single active neuron do not need to go into the recursions.
This trick alone can save a great amount of time; see Section~\ref{sec:ex2}.
Therefore, the recursive calls involve the reduced graph $\widetilde{G}$ only, which consists of the active neurons in the bogus fixed point after \som{} has converged.
The reduced graph $\widetilde{G}$ has $\widetilde{C}$ clusters, which is the number of the clusters with multiple neurons in the original network.
Another main difference between Algorithm~\ref{alg:kp} and~\ref{alg:cp} is that the argument $U$ in the \textbf{clique} function is no longer a set, it is now an aggregation of $\widetilde{C}$ sets, each of which stores the neurons in a different cluster, so that the partite feature is preserved even after the network has been reduced.

The global set $Q$ stores the current clique under consideration.
We choose on purpose that once the first clique has been found, our algorithm terminates because of efficiency considerations; see lines 10--13 and 24--26.
These lines can be safely deleted if all the cliques, i.e., matched patterns by the input probe, are required to be retrieved instead of a particular one.
At line 9, the variable \textit{level} is simply for notation convenience, so that we know the current size of $Q$.
Meanwhile, \textit{level} is not only the current cluster we are checking, but also the current \textit{level} of the binary search tree.
The maximum value for \textit{level} is bounded by $\widetilde{C}$.

In addition, we need a simple helper function \textbf{update}, which essentially does the same thing as $U\cap N(v)$ at line 17 of Algorithm~\ref{alg:cp}, updating the sub-graph of our next level recursion, adapted to the fact that $U$ is now a set of sets.

Two interesting steps in Algorithm~\ref{alg:kp} are lines 14 and 33, which are two sorting procedures.
At line 14, the nodes in the current cluster are sorted according to their degrees, so that the ones with fewer connections will be expanded first due to line 19.
At line 33, all the clusters in the next recursive level are sorted according to the active neurons they have after updating the sub-graph, so that the cluster with the fewest active neurons after the update will be the next one to expand.
A nice property of the algorithm is that, once a cluster has been determined to be the next one to expand, we do not have to sort them again (line 33 starts with \textit{level+1}, with upper levels untouched), because each time, line 19 will take out one neuron from the current cluster.
Therefore, the current cluster will remain to be the one with fewest active neurons when the algorithm returns back from a deeper level.
The purpose of these two sorting procedures is to ensure the search domain is a \emph{flat} tree instead of a \emph{deep} one.
This arrangement can bring acceleration because of two reasons:
\begin{enumerate}
	\item From an algorithmic point of view, a flat tree means if we prune once, a larger portion of the search domain can be discarded.
	\item From a programming point of view, a flat tree also means fewer recursive function calls.
\end{enumerate}

Due to the way we sort the neurons and clusters, the final clique tends to include neurons with fewer signals, (sub-messages with fewer frequencies), which appears to hurt the retrieval rate.
This is rather counter intuitive at the first glance.
However, a close second thought concludes the opposite.
We need to choose a neuron from some cluster anyway, so the same reasoning of \textsc{mf} ought to outperform \textsc{mm} applies here as well.
Therefore, our two sorting procedures not only produce a faster algorithm, but also increase retrieval rate.
Simulations in Section~\ref{sec:ex3} provide evidences to support this claim.

\section{Experiments\label{sec:experiment}}
In previous work~\cite{gripon2012nearly,yao2013gpugbnn}, the authors experiment mainly on easy scenarios, with the focus of comparative studies between \sos{} versus \som{}, which, we believe, is one of the reasons that the bogus fixed point problem was not previously uncovered.
Hence in this section, we will concentrate on difficult scenarios where the number of erased clusters and the number of stored messages are challenging for a given network.
All the simulations are performed on a 2.93GHz Intel Core 2 Duo Processor T9800 processor with 4GB memory.

Two testing scenarios are investigated.
A small network contains 8 clusters with 128 neurons each, storing 5000 randomly generated messages.
A large one contains 16 clusters with 256 neurons each, storing 40000 messages.
Both scenarios use 5000 test messages for retrieving, and three quarters of the clusters are erased, i.e., 6 for the small, and 12 for the large.
As suggested in~\cite{yao2013gpugbnn}, we set the reinforcement factor $\gamma=1$ in all cases.

\subsection{Bogus Fixed Point Problem\label{sec:ex0}}
\begin{figure}
\centering
\includegraphics[scale=.5]{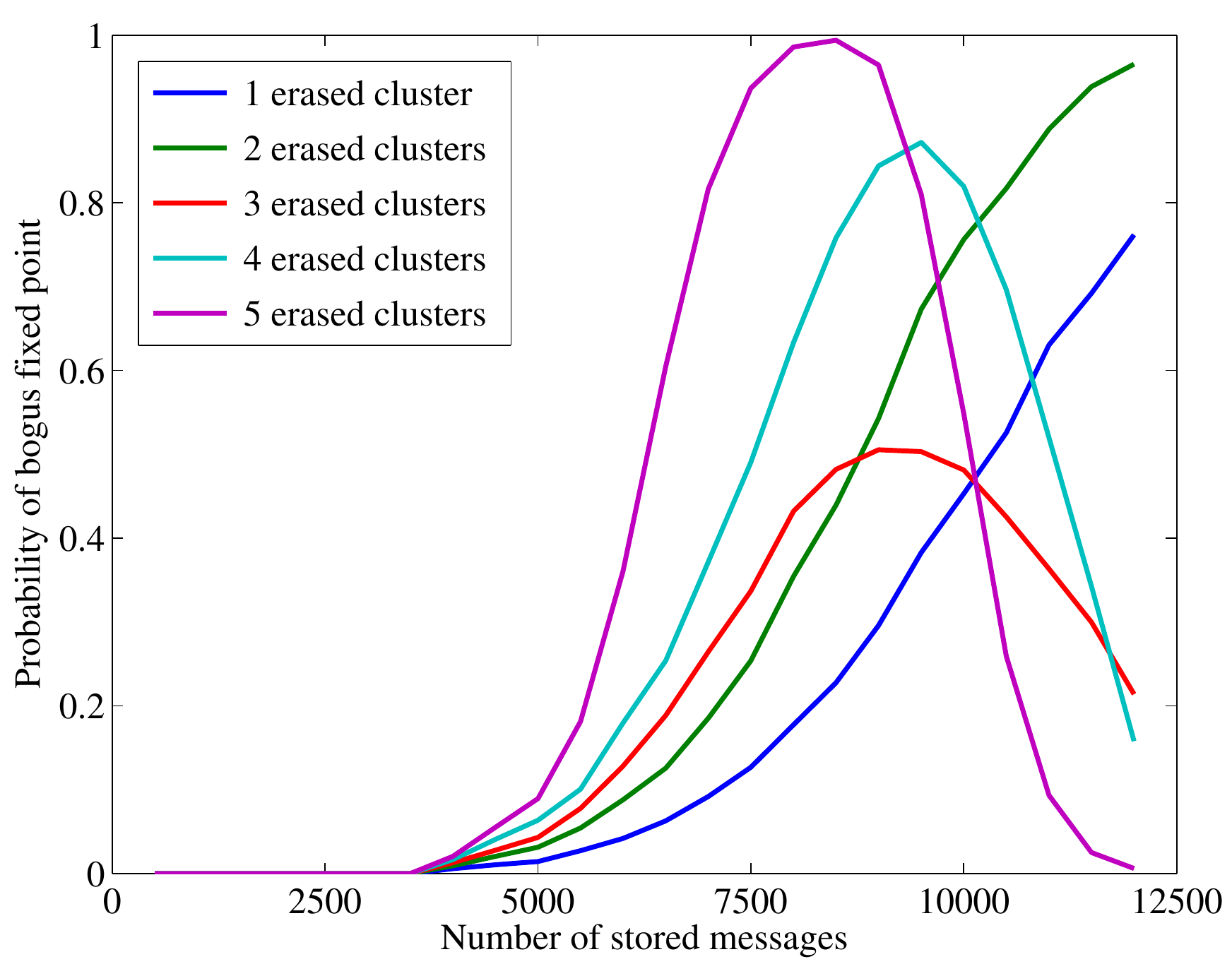}
\caption{The probability that \som{} reaches a bogus fixed point for the small scenario as the number of erased clusters and the number of stored messages increase.}
\label{fig:bogus}
\end{figure}

We have already argued that the bogus fixed point problem happens when the testing scenarios are challenging for a given network.
We plot in \fig{fig:bogus} the empirical probability that \som{} reaches a bogus fixed point for the small scenario (with number of test messages being 4000) as the number of erased clusters and the number of stored messages increase.
In order to find the probability, we first run \som{} and record the converged state as a binary vector $v_1$ with 1s being the active neurons.
Then we run any clique finding algorithm.
Instead of immediately quiting after finding the first clique, we try to find all the cliques in the converged state.
We represent the ensemble of all the cliques found as another binary vector $v_2$.
If $v_1 \neq v_2$, there must exist active neurons that do not form a clique, hence the converged state is indeed a bogus fixed point.
We increment a counter in this case, and the probability is equal to the ratio between the counter and the number of test messages.

The bogus fixed point problem happens regardless of the network size, although we only plot the small scenario for illustration purposes.
As we can see from \fig{fig:bogus}, the probability that \som{} converges to a bogus fixed point is a complicated function of both the number of erased clusters and stored messages.
When the number of erased clusters is small (e.g., 1 or 2), the probability increases with the number of stored messages, which also holds for the initial part of the curves when the number of erased clusters is large (e.g., 3, 4 or 5).
Intuitively, more stored messages saturate the network by adding more cliques.
Therefore, it is more likely for the network to converge to a bogus fixed point.
Quite interestingly, the tail of the curves decreases to 0 when the number of erased clusters is large.
This happens because too many clusters are missing and the input probe provides too little information to retrieve the desired clique.
As a result, all neurons in erased clusters remain active, which makes $v_1$ identical with $v_2$.
In this case, although the probability of bogus fixed point is small, the retrieved pattern is useless.

\subsection{Different Heuristics\label{sec:ex1}}
The approaches we test here include the original random selection scheme and six heuristics proposed in Section~\ref{sec:heuristic}.
We mentioned in Section~\ref{sec:heuristic}, the proposed heuristics do not guarantee a clique to be found.
Therefore, in our implementation we record the states along the way, so that once a cluster has no active neurons, we can rewind the state and provide the post-processing a second chance when converting the binary encoding back to the message form.
We also compare them with the clique finding algorithm in~\cite{konc2007mcqd} (\textsf{mcqd}).
It is the high retrieval rate this algorithm offers that motivated us to develop the clique finding algorithm, Algorithm~\ref{alg:kp}, which is more customized to GBNN.
Comparisons and accelerations between different clique finding algorithms are carried out in the next sub-section.

\begin{figure*}
\centering
\subfloat[]{
\includegraphics[scale=0.45]{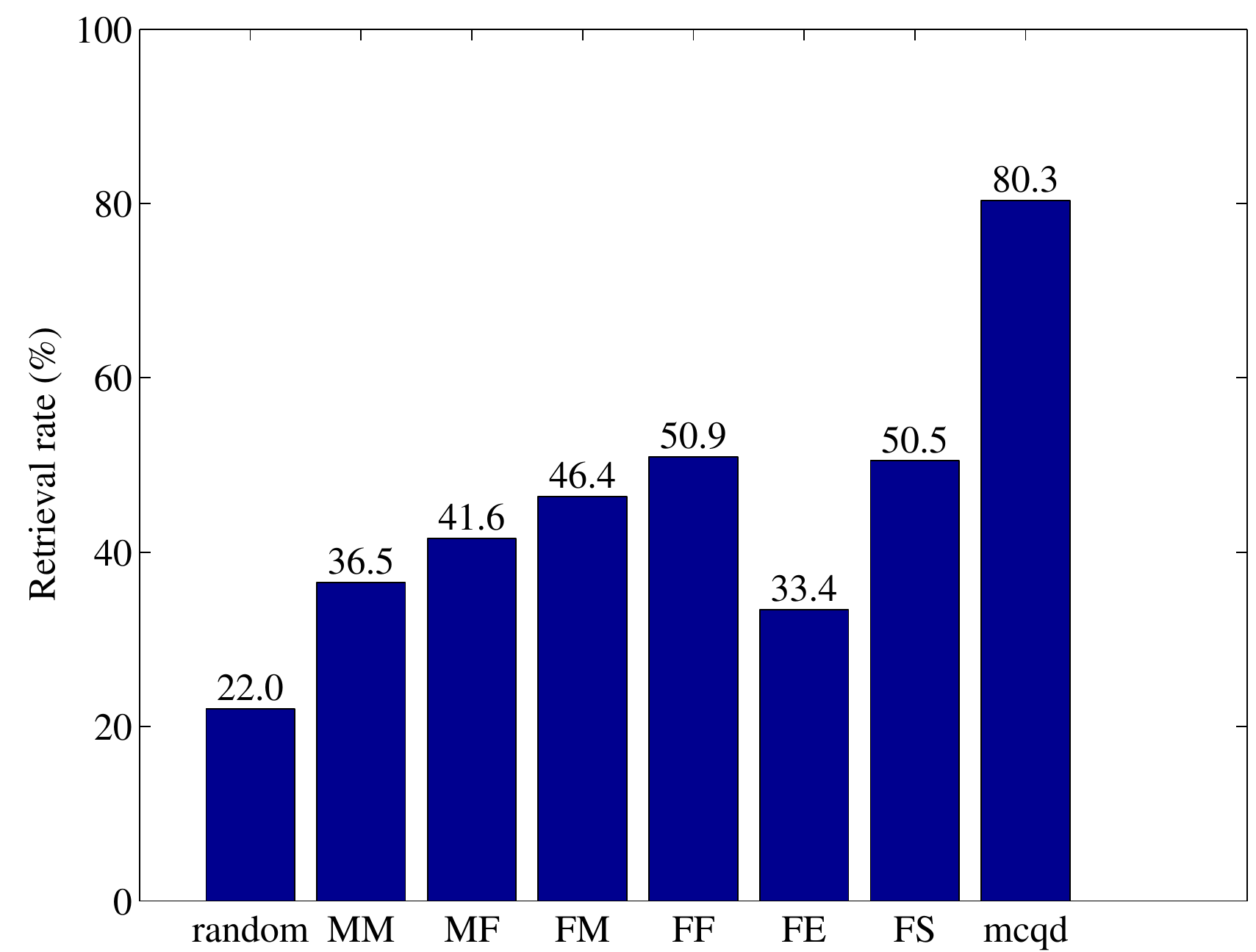}
\label{fig:ex1smallrate}
}
\subfloat[]{
\includegraphics[scale=0.45]{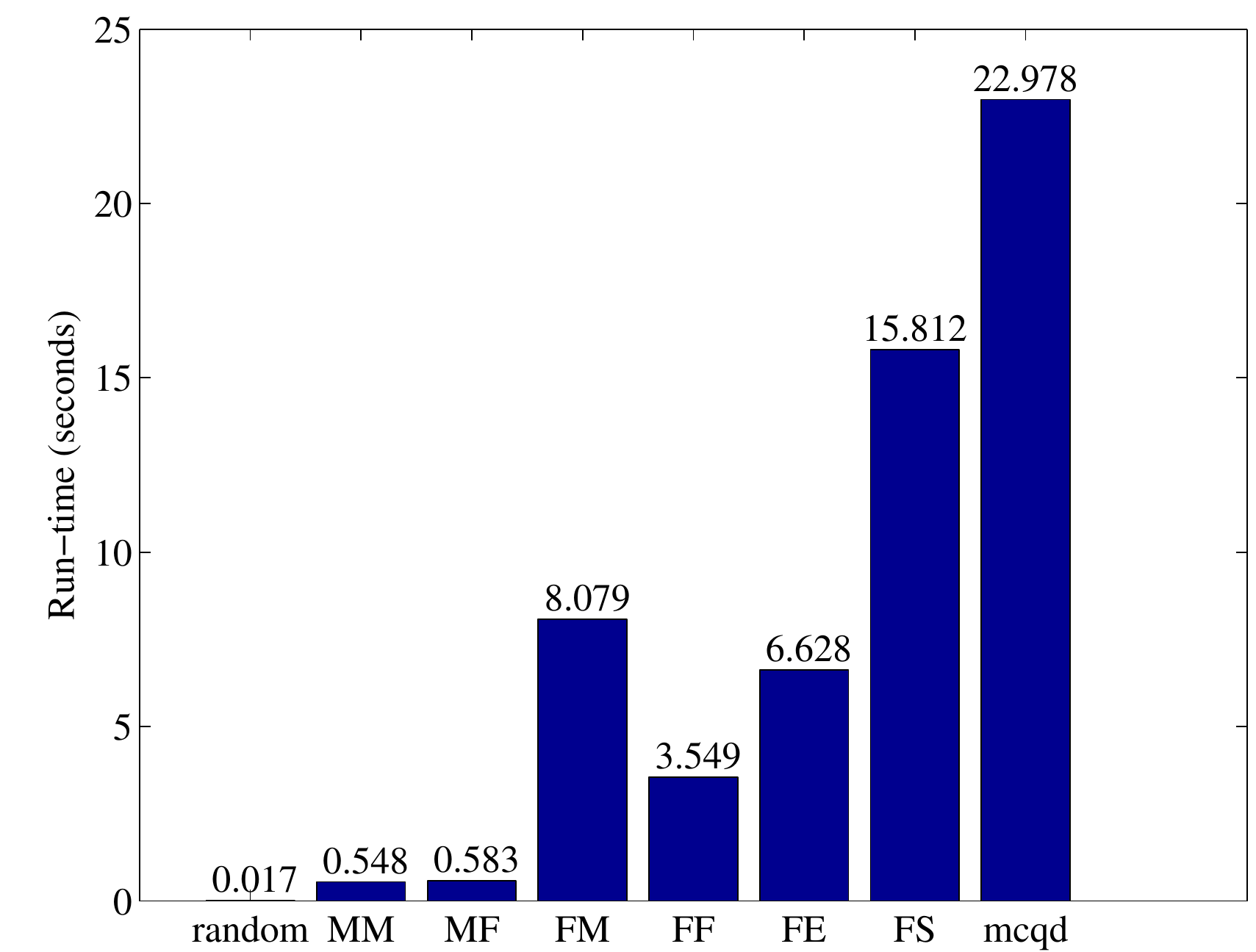}
\label{fig:ex1smalltime}
}\\
\subfloat[]{
\includegraphics[scale=0.45]{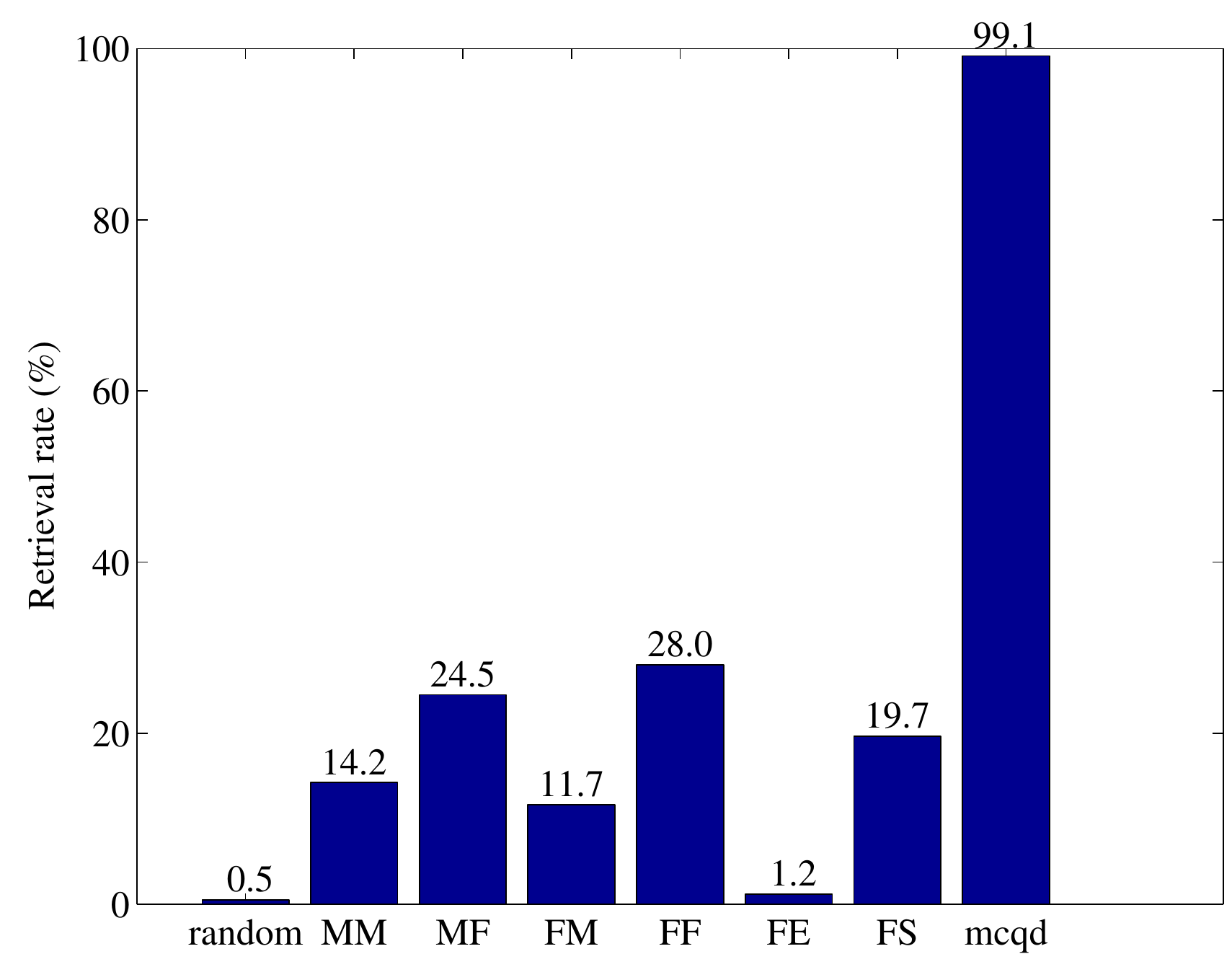}
\label{fig:ex1largerate}
}
\subfloat[]{
\includegraphics[scale=0.45]{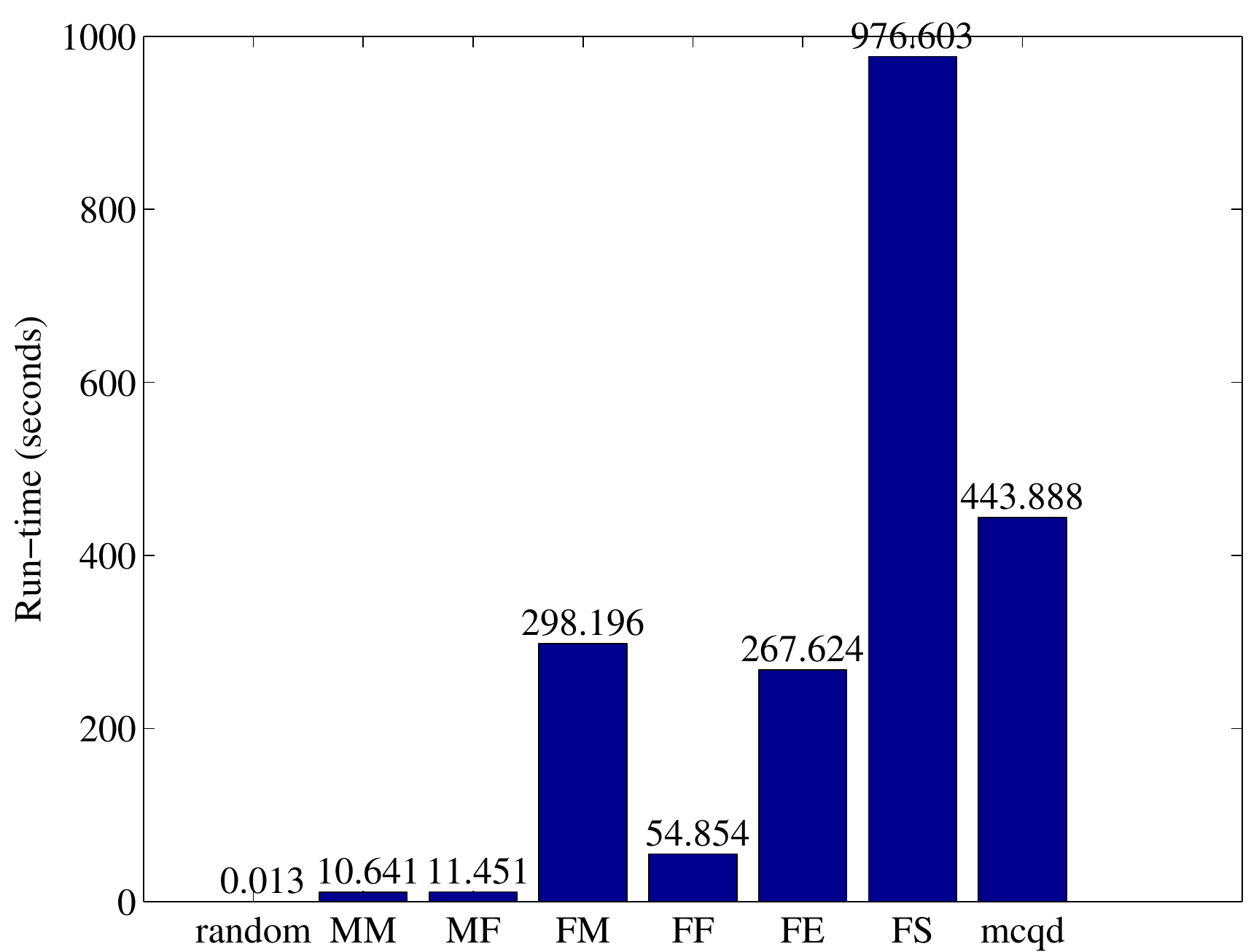}
\label{fig:ex1largetime}
}
\caption{
	Comparisons of different heuristics to escape from the bogus fixed point, where \textsf{random} is the standard \som{} implemented in prior work and \textsf{mcqd} is the algorithm in~\protect\cite{konc2007mcqd}.
	\protect\subref{fig:ex1smallrate} and \protect\subref{fig:ex1smalltime} are the retrieval rate and run-time for the small testing scenario respectively, whereas \protect\subref{fig:ex1largerate} and \protect\subref{fig:ex1largetime} are for the large scenario.
	Only the post-processing time is reported.
	Before converging, \som{} runs in about 6 seconds for the small scenario and about 100 seconds for the large scenario across the board.
	The exciting results by the clique finding approach (\textsf{mcqd}) encourage us to develop our proposed Algorithm~\ref{alg:kp} customized to GBNN's structure.
}
\label{fig:heuristic}
\end{figure*}

We report both the retrieval rate and the run-time for different heuristics in~\fig{fig:heuristic}.
The run-time reported is only for the post-processing part after \som{} has converged to the bogus fixed point.
We see from~\fig{fig:ex1smallrate} and~\fig{fig:ex1largerate} that all of the heuristics are better than the random selection scheme for the standard \som{} in terms of retrieval rate.
In Section~\ref{sec:heuristic}, we remarked that \textsc{fm} and \textsc{ff} ought to outperform \textsc{mm} and \textsc{mf}.
This is true for the small scenario, but it does not hold for the large case (\textsc{fm} is much worse than the rest).
We also argued in Section~\ref{sec:heuristic} that \textsc{mf} ought to be better than \textsc{mm}, which is evident for both scenarios.
Although \textsc{fe} tries to reflect the pattern frequencies, it does not work out for either case.
This is mainly because we concentrate on difficult situations where the number of the stored patterns are demanding for a given network, the resulting GBNN is thus highly connected.
The frequencies are no longer a valid indicator (imagine the extreme case, when every neuron connects to every other neurons in different clusters).
\textsc{fs} performs better than \textsc{fe} since it is only interested in the subset of neurons addressed by the probe and a lot of unnecessary edges are eliminated before it makes the decision.
The clique finding approach is quite encouraging in both scenarios: for the small case, the performance quadruples from 20\% to 80\%, and it almost perfectly recovers all the queries in the large setting.

In terms of run-time, from~\fig{fig:ex1smalltime} and~\fig{fig:ex1largetime}, we can tell that \textsc{mm} and \textsc{mf} are much faster than the rest, which accords with our judgement in Section~\ref{sec:heuristic}.
The \textsc{fs} heuristic is really slow --- even slower than the clique finding approach in the large scenario.
Considering both retrieval rate and run-time, we argue that \textsc{ff} makes a better balance than any other heuristics between these two metrics.

The astonishing retrieval rate but slow run-time of the clique finding approach (\textsf{mcqd}) indeed stimulates us to work on faster solutions. 

\subsection{Different Clique Finding Algorithms\label{sec:ex2}}
In the previous sub-section, we see terrific performance gain from exploiting clique finding algorithms.
However, the run-time is not satisfying.
This sub-section deals with the problem of comparing and accelerating the clique finding approaches, so that we can have both correctness and speed.

\begin{figure}
\centering
\includegraphics[scale=0.5]{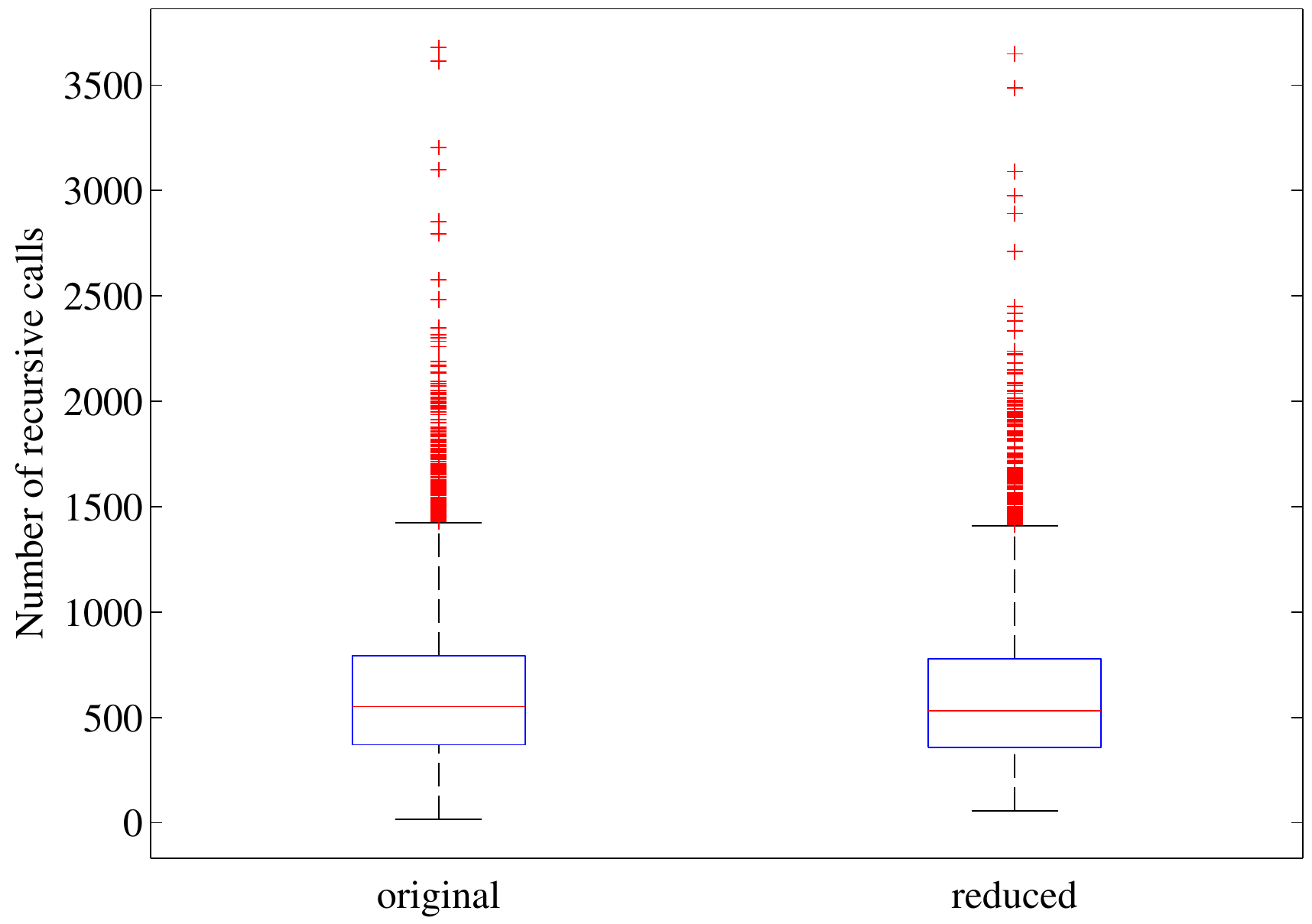}
\caption{
	Comparisons of \textsf{mcqd} for the large scenario between the original and the reduced graph in terms of the number of recursive function calls.
}
\label{fig:reduce}
\end{figure}

We will focus on three algorithms, i.e., the fastest variant available of the vertex coloring approach~\cite{konc2007mcqd}, the classic~\cite{carraghan1990exact} as in Algorithm~\ref{alg:cp} and our newly developed Algorithm~\ref{alg:kp}.
We first demonstrate that renumbering and reordering the original GBNN into a reduced graph brings tremendous acceleration.
\fig{fig:reduce} shows the box plot of the number of recursive function calls, the original versus the redueced graph, for the large setting using \textsf{mcqd}.
We cannot tell a significant disagreement between these two graphs in terms of the number of recursive calls.
However, the run-time is a totally different story.
The original graph requires 439.011 seconds, whereas the reduced graph only needs 6.197 seconds, which indicates that the reduced graph will cut the time in each level of the recursive calls.

\begin{figure*}
\centering
\subfloat[]{
\includegraphics[scale=0.45]{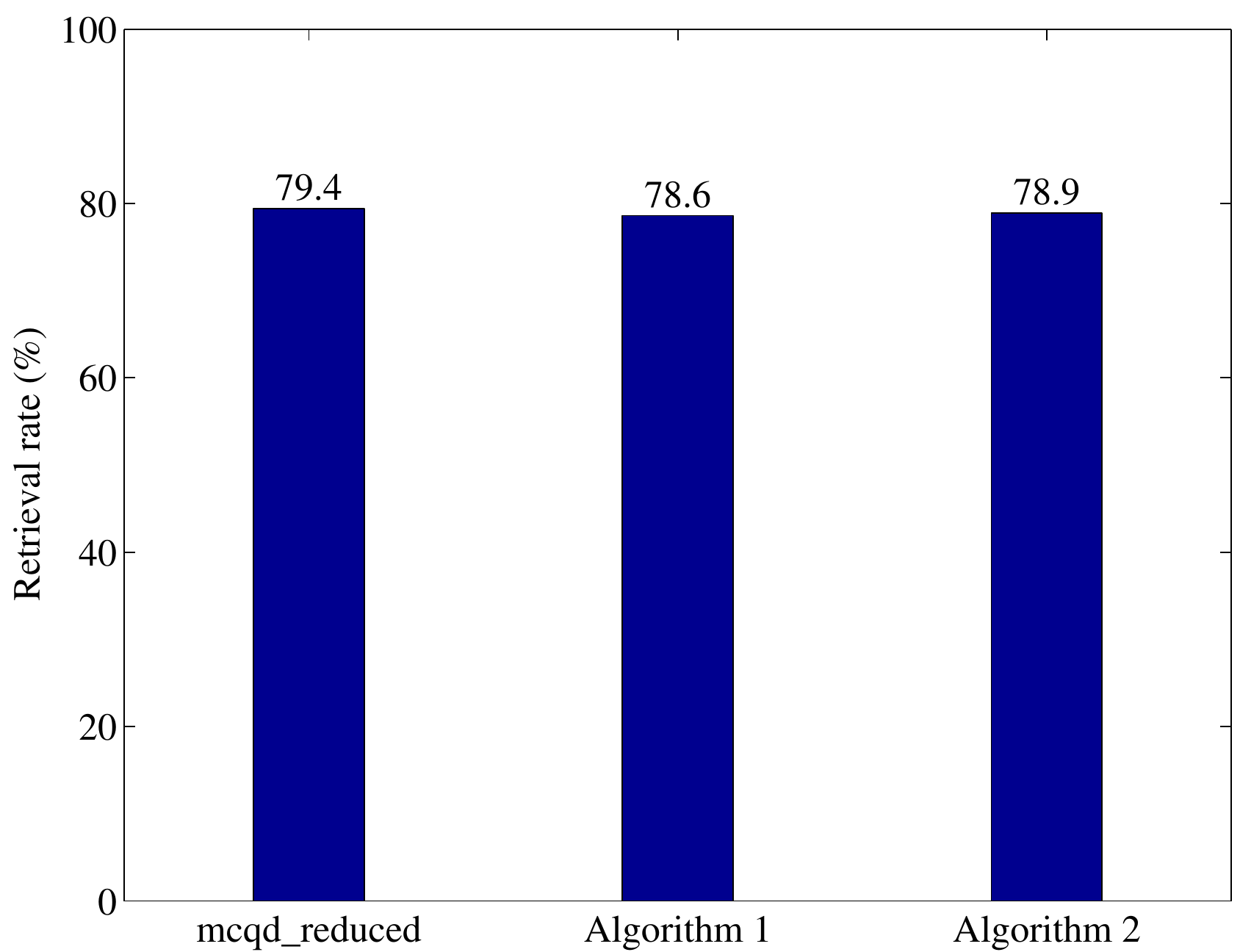}
\label{fig:ex2smallrate}
}
\subfloat[]{
\includegraphics[scale=0.45]{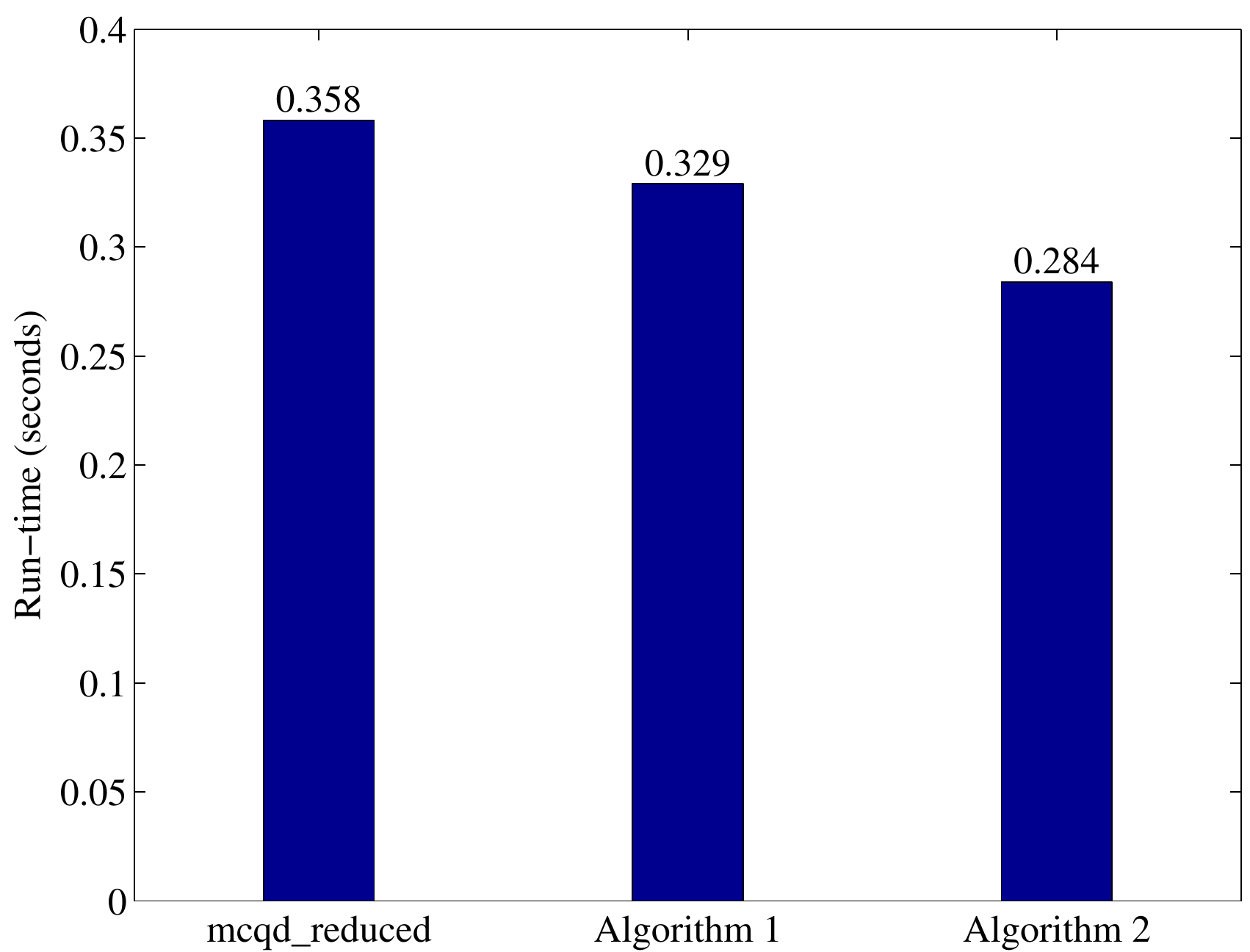}
\label{fig:ex2smalltime}
}\\
\subfloat[]{
\includegraphics[scale=0.45]{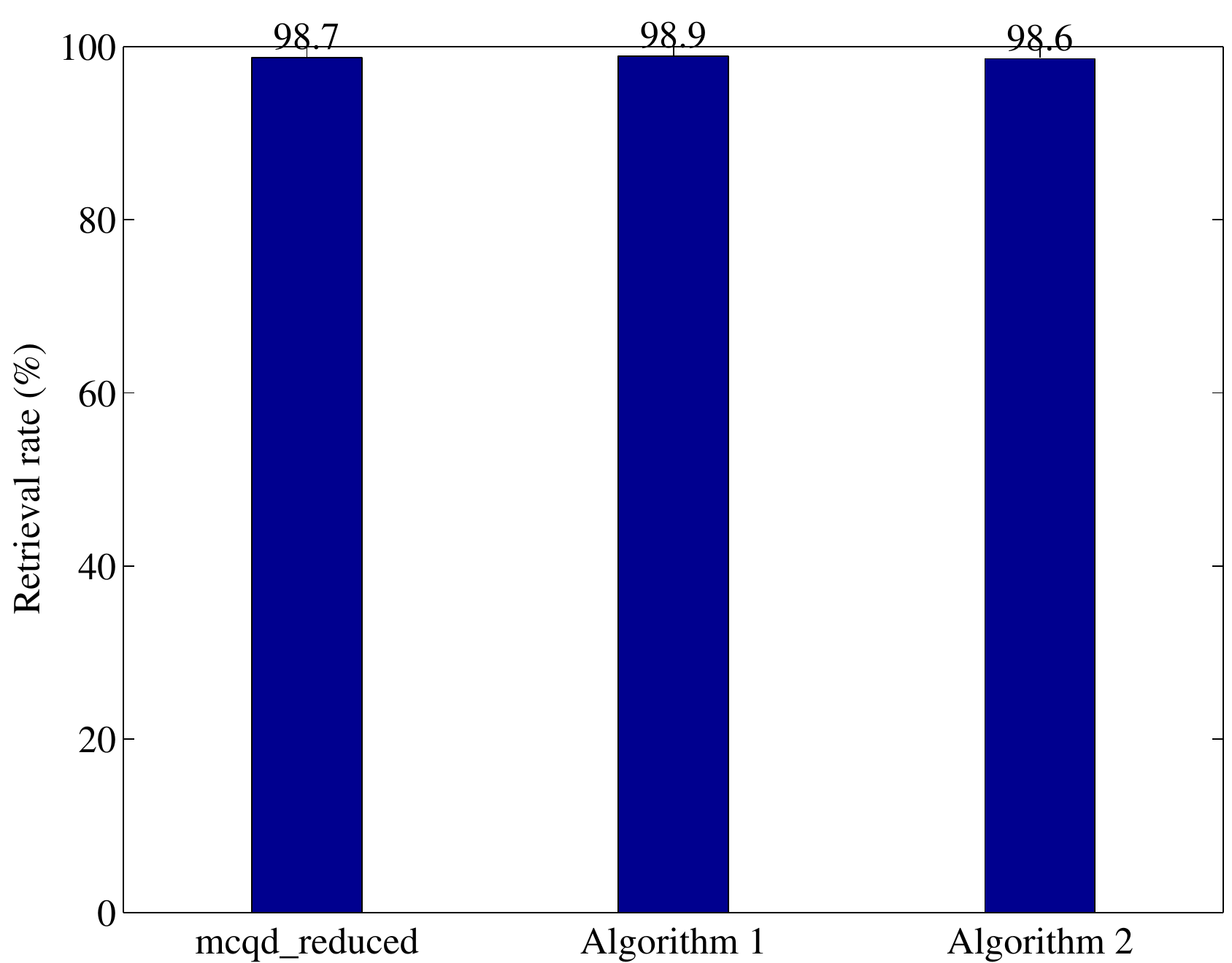}
\label{fig:ex2largerate}
}
\subfloat[]{
\includegraphics[scale=0.45]{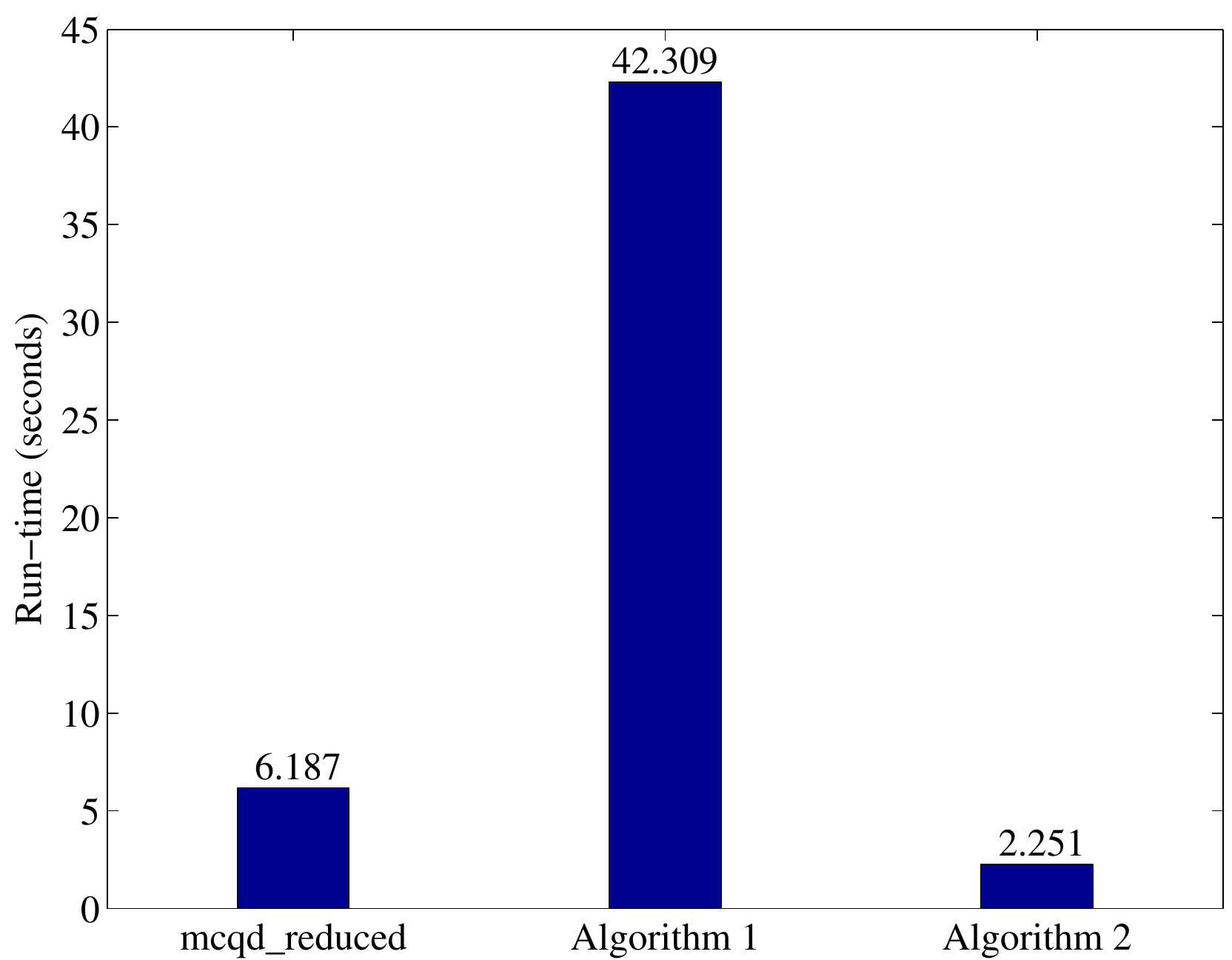}
\label{fig:ex2largetime}
}
\caption{
	Comparisons of different clique finding algorithms in different experiment settings.
	\protect\subref{fig:ex2smallrate} and \protect\subref{fig:ex2smalltime} are the retrieval rate and run-time for the small scenario respectively, whereas \protect\subref{fig:ex2largerate} and \protect\subref{fig:ex2largetime} are for the large setting.
}
\label{fig:cliquecompare}
\end{figure*}

Then we apply the same reducing trick to all the algorithms in comparison.
\fig{fig:cliquecompare} presents the retrieval rate and run-time for different clique finding algorithms in both small and large settings.
All of the retrieval rates are identical and promising for different algorithms, which also validates that our implementations are correct.
The run-time of the newly developed Algorithm~\ref{alg:kp} is a big win over all the other alternatives, not only among the clique finding algorithms, but also across all the other heuristics; see~\fig{fig:heuristic}.

\subsection{Sorting Procedures in Algorithm~\protect\ref{alg:kp}\label{sec:ex3}}
First we would like to provide evidence that compared to a deep search tree, a flat one not only accelerates the algorithm but also brings us a better retrieval rate.
To make such an argument, we first run the newly developed Algorithm~\ref{alg:kp}, and then change both sorting procedures in the reversed order so that a deep search tree will be constructed.
For a better illustration, we take the large scenario again and increase the number of stored messages to be 50000 to challenge the algorithms even further.
The flat tree runs in 42.573 seconds, giving us 77.4\% of successful retrievals, whereas the deep tree runs in 584.337 seconds, giving us 65.6\% successful retrievals.
This is a 14$\times$ faster version with over 10\% performance gain using almost the same algorithm except for the branching order.

\begin{figure}
\centering
\includegraphics[scale=.5]{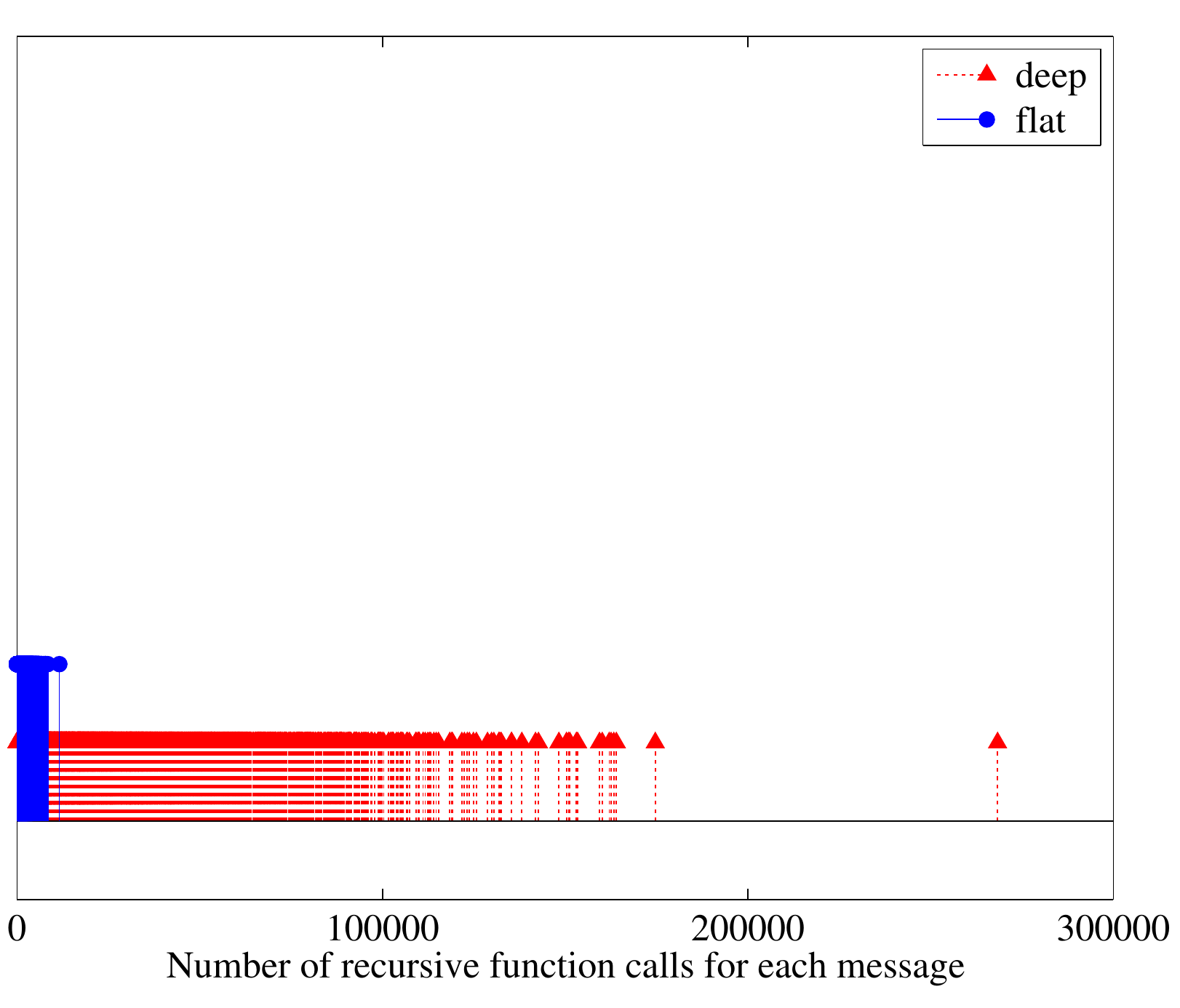}
\caption{The number of recursive function calls for each message by the flat and the deep tree approach respectively. The red dashline triangles are for the deep tree while the blue circles are for the flat tree. The flat tree needs much less function calls.}
\label{fig:flatdeep}
\end{figure}

Also see~\fig{fig:flatdeep}, which shows the number of recursive calls either approach goes through to retrieve a message.
The red dashline triangles are for the deep tree, and the blue circles are for the flat tree.
We can easily tell that to retrieve some messages, the deep tree approach even has to invoke nearly 300000 functions calls, while in general the flat tree requires much fewer function calls than the deep one.
To be more precise, the medians of the number of function calls for the flat and deep trees are 1008 and 19082 respectively, which is roughly 20$\times$ smaller for the flat case.

Finally, we would like to know if these two sorting procedures contribute equally to the algorithm.
We first run the algorithm with both sorting turned on.
The algorithm runs in 43.072 seconds with retrieval rate of 77.4\%.
Then we only keep on the sorting within the current cluster according to the node degrees (line 14 of Algorithm~\ref{alg:kp}).
It runs in 106.541 seconds with the retrieval rate unchanged.
We again run the algorithm, this time with the sorting between clusters turned on (line 33 of Algorithm~\ref{alg:kp}).
It runs in 42.595 seconds with the retrieval rate dropping to 71.1\%.
All three versions run against the same data, eliminating all undesired factors.
The results indicate that the order of the expanded cluster is crucial to the correctness of the decoding process.
Once the cluster order has been determined, which node to branch first mainly affects the run-time.

\section{Summary\label{sec:summary}}
GBNN is a recently invented recurrent neural network embracing LDPC-like sparse encoding setup, which makes it extremely resilient to noise and errors.
Two activation rules exist for the activation of the neuron dynamics, namely \sos{} and \som{}.
In this work, we look into the activation rules themselves.
\sos{} focuses on individual signals whereas \som{} concentrates on cluster-wise signal contributions.
This particular trait helps \som{} to stand out in terms of successful retrieval rate by a large margin.
However, the same peculiarity ensnares \som{} when it has already reached the converged state.
We identify such an overlooked situation for the first time and propose a number of heuristics to facilitate \som{}'s decoding process, pushing it beyond the bogus fixed point, by taking into account the individual signals which was \sos{}'s original spirit.
Prior work, e.g.,~\cite{yao2013gpugbnn}, combines \sos{} and \som{} mainly due to computational considerations, so that \sos{} can be exploited to accelerate the \som{} process, whereas this work blends these two in a totally orthogonal perspective.

To solve the bogus fixed point problem directly and completely, a post-processing algorithm is also developed, which is to find a maximum clique in the network essentially.
The algorithm is tailored to accommodate the special property of GBNN being a $C$-partite graph.
Experiment results show that the algorithm outperforms the random selection of the standard \som{} scheme and all the heuristics proposed in this work, in terms of both retrieval rate and run-time, which also suggest that there is plenty of room to improve the activation rules themselves.

Possible directions of future research may include stochastic activation schemes which has been done in Boltzmann machines or deep learning networks, so that the retrieval process for GBNN does not need to be divided into two distinct stages.
It is also a good heuristic to introduce a heat parameter as in Boltzmann machines in the decoding process, since it might avoid the plateau that corresponds to the bogus fixed point of \som{}.
In addition, we are interested in adapting and testing \sos{} or \som{} in general networks other than partite graphs, e.g., Erd\H{o}s-R\'{e}nyi graphs.

\section*{Acknowledgement}
\addcontentsline{toc}{section}{Acknowledgement}
This work was funded, in part, by the Natural Sciences and Engineering Research Council of Canada (NSERC), the \emph{Fonds Qu\'{e}b\'{e}cois de la recherche sur la nature et les technologies} (FQRNT) and the European Research Council project NEUCOD.
\bibliographystyle{IEEEtran}
\bibliography{my}

\end{document}